\documentclass{article}
\pdfoutput=1

\usepackage{PRIMEarxiv}

\usepackage{threeparttable}
\usepackage{subfig}
\usepackage[utf8]{inputenc} % allow utf-8 input
\usepackage[T1]{fontenc}    % use 8-bit T1 fonts
\usepackage{hyperref}       % hyperlinks
\usepackage{url}            % simple URL typesetting
\usepackage{booktabs}       % professional-quality tables
\usepackage{amsfonts}       % blackboard math symbols
\usepackage{nicefrac}       % compact symbols for 1/2, etc.
\usepackage{microtype}      % microtypography
\usepackage{lipsum}
\usepackage{fancyhdr}       % header
\usepackage{graphicx}       % graphics
\graphicspath{{media/}}     % organize your images and other figures under media/ folder
\title{Identification and classification of exfoliated graphene flakes from microscopy images using a hierarchical deep convolutional neural network
}

\author{
 Soroush Mahjoubi \\
  Department of Civil, Environmental and Ocean Engineering\\
  Stevens Institute of Technology\\
  Hoboken, New Jersey 07030, USA \\
   \And
 Fan Ye \\
  Department of Electrical and Computer Engineering\\
  University of Massachusetts Amherst\\
  Amherst, Massachusetts 01003, USA \\
  \And
 Yi Bao \\
  Department of Civil, Environmental and Ocean Engineering\\
  Stevens Institute of Technology\\
  Hoboken, New Jersey 07030, USA \\
  \texttt{yi.bao@stevens.edu} \\
    \And
Weina Meng \\
  Department of Civil, Environmental and Ocean Engineering\\
  Stevens Institute of Technology\\
  Hoboken, New Jersey 07030, USA \\
  \texttt{weina.meng@stevens.edu} \\
    \And
 Xian Zhang \\
  Department of Mechanical Engineering\\
  Stevens Institute of Technology\\
  Hoboken, New Jersey 07030, USA \\
  \texttt{xian.zhang@stevens.edu} \\
}

\begin{document}
\maketitle

\begin{abstract}
Identification of the mechanically exfoliated graphene flakes and classification of the thickness is important in the nanomanufacturing of next-generation materials and devices that overcome the bottleneck of Moore’s Law. Currently, identification and classification of exfoliated graphene flakes are conducted by human via inspecting the optical microscope images. The existing state-of-the-art automatic identification by machine learning is not able to accommodate images with different backgrounds while different backgrounds are unavoidable in experiments. This paper presents a deep learning method to automatically identify and classify the thickness of exfoliated graphene flakes on Si/SiO2 substrates from optical microscope images with various settings and background colors. The presented method uses a hierarchical deep convolutional neural network that is capable of learning new images while preserving the knowledge from previous images. The deep learning model was trained and used to classify exfoliated graphene flakes into monolayer (1L), bi-layer (2L), tri-layer (3L), four-to-six-layer (4-6L), seven-to-ten-layer (7-10L), and bulk categories. Compared with existing machine learning methods, the presented method possesses high accuracy and efficiency as well as robustness to the backgrounds and resolutions of images. The results indicated that our deep learning model has accuracy as high as 99\% in identifying and classifying exfoliated graphene flakes. This research will shed light on scaled-up manufacturing and characterization of graphene for advanced materials and devices.

\end{abstract}

\section{Introduction}
Two-dimensional (2D) materials, endowed with superior electrical [1], ferromagnetic [2, 3], semiconducting [4], superconducting [5], thermal [6-10], and optical properties [11, 12], are the next-generation materials in the areas of mechanical, photonic, and electronic. 2D materials are widely obtained by mechanical exfoliation using scotch tapes [13, 14], which ensures the most pristine properties and conditions of 2D materials. Mechanical exfoliation methods yield flakes of 2D materials with random thickness, shapes, and morphology on the substrate, followed by a manual search under an optical microscope by human intuitions. For scientific and device development purposes, the ideal flakes have a uniform thickness. Many interesting phenomena only occur at samples of one to a few atomic layers. In current practices, flake thickness is mainly evaluated by human operators through examination of the color and contrast of flakes from the microscope images. The flakes are then further evaluated by material characterization equipment, such as Raman microscope and atomic force microscope [15-20]. There are several disadvantages in this process: manual examination inevitably yields errors due to different colors and background conditions regarding color and brightness; and characterization of 2D flakes is costly, inefficient, and time-consuming. Therefore, an efficient and accurate method for identifying the thickness is highly desired [21-24].

A few automatic methods to identify 2D flakes from optical microscope images were reported in references [17, 25-29]. Five major limitations were identified from the existing automatic methods: (1) The existing methods were designed for specific microscope conditions such as background color and light brightness, compromising the applicability for different conditions; (2) the existing machine learning models are not capable of learning new micrographs, so re-training the existing models using new data leads to catastrophic loss of the knowledge learned from old data [30]; (3) optical microscope images often suffer from artifacts such as poor contrast, vignetting, and overexposure. Multiple methods were developed to enhance microscope images for human vision [31, 32], but not applicable for machine vision; (4) imbalanced class distribution in training images compromises the evaluation of under-represented classes [33]. The areas of the thin flakes are relatively small compared with the areas of the substrate (background) and bulk flakes. (5) The classification precision of graphene thickness was rough. The flakes were categorized into three classes [25], which are insufficient in many applications.

In this work, we present a novel hierarchical deep learning method based on an unsupervised classification model and multiple semantic segmentation models and evaluate the performance of the method in automatically identifying exfoliated graphene flakes and determining thicknesses from optical microscope images. The flakes are categorized into six classes: monolayer (1L), bi-layer (2L), tri-layer (3L), four-to-six-layer (4-6L), seven-to-ten-layer (7-10L), and bulk. Novel computer vision techniques are presented to improve microscope images using an image quality metric and improved adaptive gamma correction [34]. The proposed method leverages weak learning, data augmentation, iterative stratification [35], and weighted cross-entropy loss to improve the accuracy and the generalization performance. The effects of the resolution and the background condition of images are considered to improve the robustness of the deep learning model. This method promotes the capability of processing and manufacturing 2D materials and devices by improving efficiency and accuracy while minimizing human intervention.

\section{Methodology}
\label{sec:headings}

We produced 2D flakes using the scotch tape mechanical exfoliation method [13, 14] from bulk crystals (Graphene Supermarket). After bulk crystal was folded on scotch tape for multiple times, 2D flakes were directly transferred onto the SiO2/Si substrate by a 2-minute scratch. The SiO2 thickness was 285 nm for the best optical contrast to identify the thickness of 2D flakes. An optical microscope (Nikon Eclipse LV150N) was used to capture RGB images of graphene flakes from 1L up to 10 L on the SiO2/Si substrate using different microscope settings. A total of 273 high-resolution optical microscope images were captured. An atomic force microscope (Bruker BioScope Resolve) and Raman spectroscope (Renishaw inVia Confocal) were used to evaluate the thickness of each flake. Pixel-wise ground truth labels of MATLAB were annotated with the thickness evaluation results. The pixels were categorized into seven classes: background, 1L, 2L, 3L, 4-6L, 7-10L, and bulk.

\begin{figure}
  \centering
  \includegraphics[width=14cm]{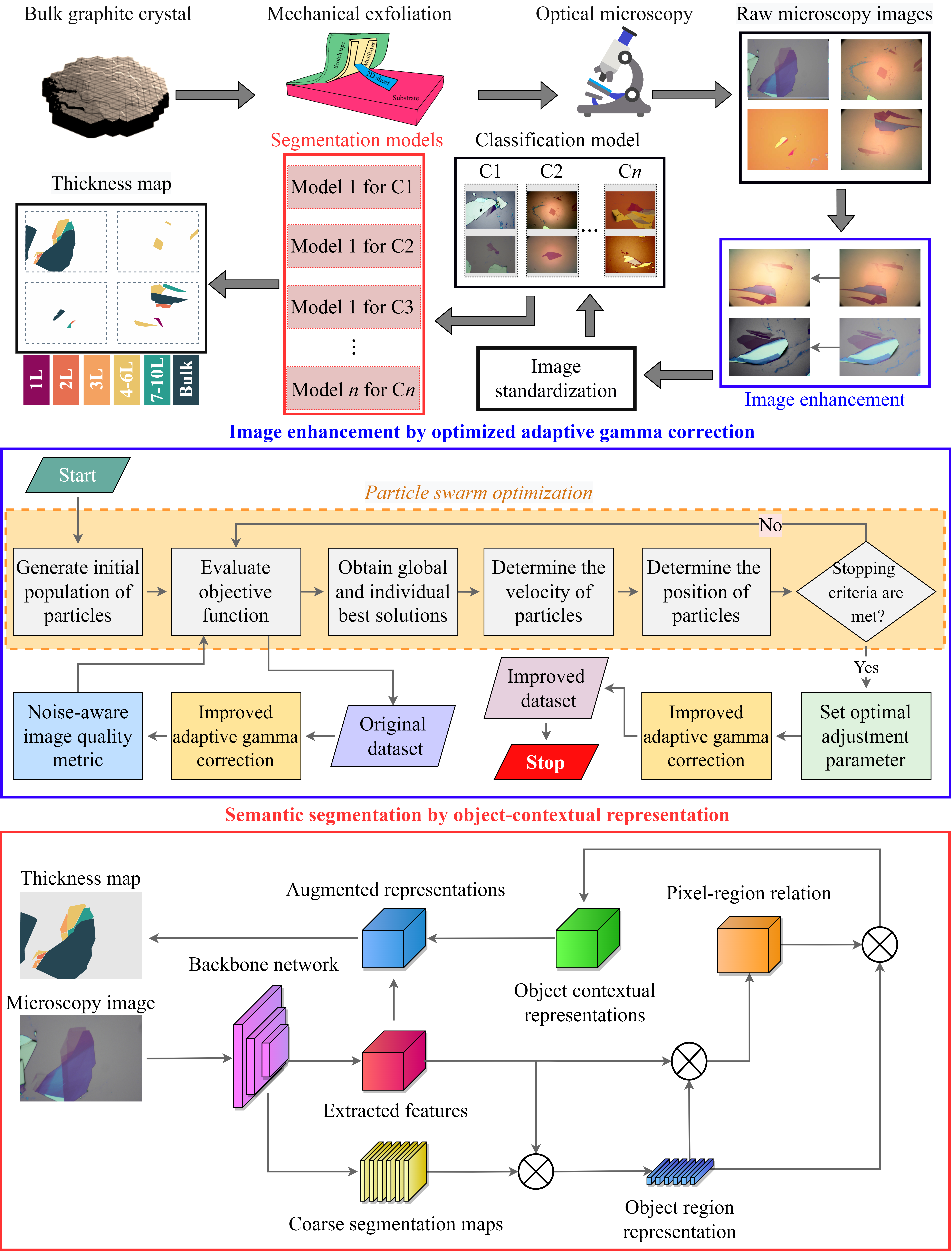}
  \caption{Flowchart of the robust machine intelligence framework for classification of 2D flakes.}
  \label{fig:fig1}
\end{figure}

Figure \ref{fig:fig1} shows the flowchart with five main steps: (1) Dataset development: Optical microscopy images are captured to develop a dataset for training a machine learning model (Section 2.1). (2) Dataset enhancement: The quality of images is improved using multiple methods developed in this study (Section 2.2). (3) Dataset standardization and categorization: The images are standardized and categorized using an unsupervised machine learning model (Section 2.3). (4) Semantic segmentation: A deep convolutional neural network is trained to identify exfoliated graphene flakes and quantify layer numbers (Section 2.4). (5) Performance evaluation: The performance of the framework is evaluated using performance metrics (Section 2.5).

\subsection{Statistical Analysis}

\begin{figure}%
    \centering
    \subfloat[\centering]{{\includegraphics[width=7.5cm]{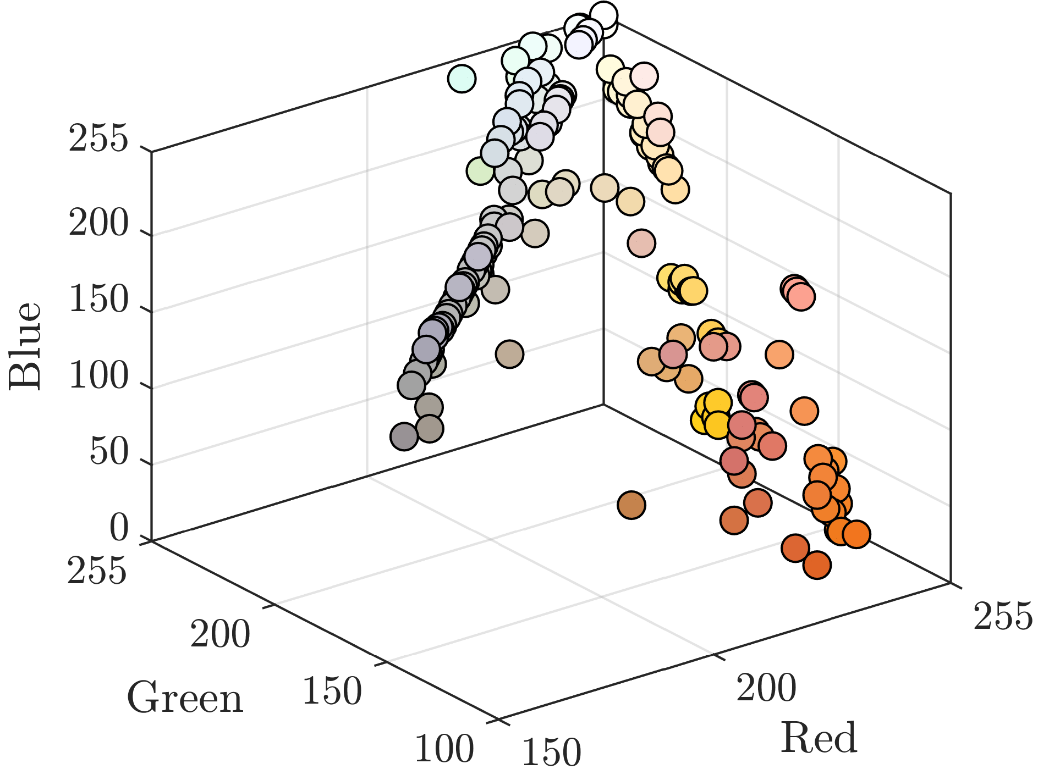} }}%
    \qquad
    \subfloat[\centering]{{\includegraphics[width=7.5cm]{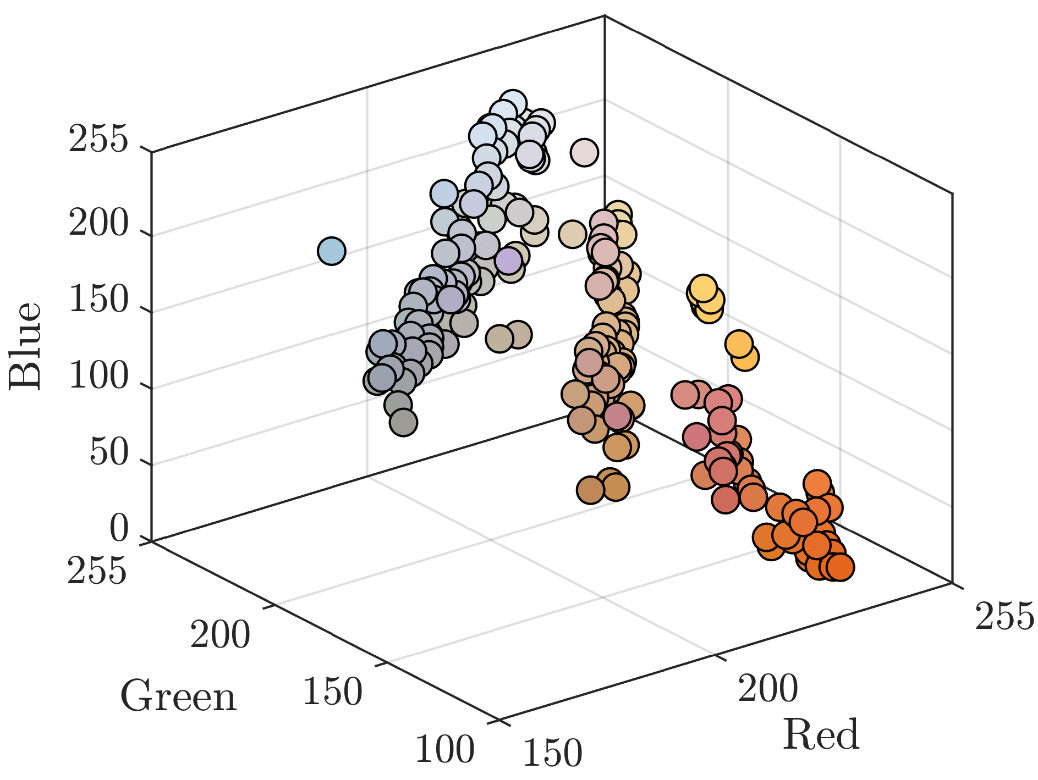} }}%
    \caption{Scatter plots of pixel intensity levels: (a) the pixel intensity of the background, and (b) the average pixel intensity of images. Each dot represents a microscope image. The color of dot represents the calculated color value.}%
    \label{fig:fig2}
\end{figure}

Statistical analysis was conducted using the ground truth labels to evaluate the distributions of the seven classes. The weight of the i-th class was defined as the number of pixels of the i-th class divided by the total number of pixels. Table~\ref{tab:table1} lists the results of the mean, median, and maximum class weights of the seven classes, and the percentage of images with a class zero-class weight. The mean class weights of classes 1, 3, 4, and 5 were less than 1\%. The median class weights of classes 1 to 6 were zero, while the median class weight of background was higher than 95\%. The percentages of images with zero class weight for classes 1 (1L) and 5 (7-10L) were higher than 75\%. The statistical results indicated that the different classes in the dataset were highly imbalanced. The interested classes 1 to 6 were under-represented. The background class was over-represented. The statistical analysis was extended to the colors of images. It was found that different images had different background colors and brightness (see Figure \ref{fig:fig2}). Due to the variation of pixel color and intensity, it is unsuitable to use color-based segmentation methods [17]. In summary, the dataset of images introduces challenges due to the imbalanced classes and scattering of background colors and brightness. 

In addition, microscopy images introduce challenges due to overexposure, specular reflection, vignetting, dirt, and out-of-focus errors: (1) Overexposure: When excessive light is received by the microscope, overexposure may occur and cause loss of detail in images. (2) Specular reflection: Overhead lighting may cause specular reflection of graphene flakes and lead to over-saturation and loss of detail. (3) Vignetting: Vignetting appears as a radial darkening towards the corners of an image. (4) Dirt: Dirt and debris cause blur images. (5) Out-of-focus: Improper focus of the microscope causes blurred images and loss of details.

\begin{table}
 \caption{Statistics of class weights}
  \centering
  \begin{threeparttable}
  \begin{tabular}{llllll}
    \toprule
Designation & Description  & Mean class & Median class & Maximum class & Images with zero  \\
of class & of class & weight & weight & weight & class weight \\
    \midrule
0 & Background & 91.99\% & 95.87\% & 99.83\% & 0.00\% \\ 
1 & $1L^1$ & 0.59\% & 0.00\% & 19.72\% & 75.63\% \\ 
2 & 2L & 1.77\% & 0.00\% & 62.62\% & 55.51\% \\ 
3 & 3L & 0.70\% & 0.00\% & 15.46\% & 55.15\% \\ 
4 & 4-6L & 0.68\% & 0.00\% & 12.17\% & 61.76\% \\ 
5 & 7-10L & 0.47\% & 0.00\% & 27.51\% & 79.04\% \\  
6 & Bulk & 3.81\% & 0.00\% & 40.61\% & 51.84\% \\ 
    \bottomrule
  \end{tabular}
  \begin{tablenotes}
  \item[1] “L” stands for layer(s).
  \end{tablenotes}
  \end{threeparttable}
  \label{tab:table1}

\end{table}

\subsection{Optimized adaptive gamma correction}

To improve the quality of images, we present an optimized adaptive gamma correction method (refer Figure \ref{fig:fig1}), through integrating the improved adaptive gamma correction method [34], noise-aware image quality metrics [38], and particle swarm optimization [37]. The development and integration of the methods are elaborated in the following subsections.

\subsubsection{Improved adaptive gamma correction}
The improved adaptive gamma correction method modifies light intensity while preserving the color properties of images with overexposure problems in the following eight steps [34]: 

\begin{enumerate}

\item RGB images are converted to Y $C_b$ $C_r$ color space, where Y is the luma component for light intensity; $C_b$ and $C_r$ are the blue-difference and red-difference chroma component, respectively. 
\item The luma component is used to convert an image into a negative image: 
\begin{equation}
{Y\left(x,y\right)}_N=255-Y\left(x,y\right)
\label{eq:eq1}
\end{equation}
where $Y\left(x,y\right)_N$ is the negative image; $Y\left(x,y\right)$ is the luma component of the input image; $x$ and $y$ are the coordinates of images: $x=1,2,\ldots,M$, and $y=1,2,\ldots,N$, where $M$ and $N$ are the width and height of an image in pixel.

\item The probability density function of pixels with an intensity level $l$ is expressed as:
\begin{equation}
PDF\left(l\right)=\frac{n_l}{MN}
\label{eq:eq2}
\end{equation}
where $n_l$ is the number of pixels with an intensity level $l$.

\item The weighting distribution function is expressed as:
\begin{equation}
W\left(l\right)={PDF}_{max}\left[\frac{PDF\left(l\right)-{PDF}_{min}}{{PDF}_{max}-{PDF}_{min}}\right]^\alpha
\label{eq:eq3}
\end{equation}
where ${PDF}_{max}$ and ${PDF}_{min}$ are the maximum and minimum probability density functions of pixels with an intensity level l, respectively; $\alpha$ is the adjusted parameter; and $l_{max}$ is the maximum intensity level of the luma component. 

\item The cumulative distribution function is expressed as:
\begin{equation}
CDF\left(l\right)=\frac{\sum_{i=1}^{l}W\left(i\right)}{\sum_{i=1}^{l_{max}}W\left(i\right)}
\label{eq:eq4}
\end{equation}

\item The modified luma component of the negative image is obtained from the following transformation function:
\begin{equation}
{I^\prime\left(x,y\right)}_N=l_{max}\left(\frac{l}{l_{max}}\right)^{1-CDF\left(l\right)}
\label{eq:eq5}
\end{equation}

\item The modified luma component of the input image is determined as:

\begin{equation}
Y^\prime\left(x,y\right)=255-{Y^\prime\left(x,y\right)}_N
\label{eq:eq6}
\end{equation}

\item The modified luma component and chroma components of the image are converted to a RGB image, as the improved image. 

\end{enumerate}

\subsubsection{Noise-aware image quality metric}
A noise-aware image quality metric was used to assess image quality based on the gradient, entropy, and noise of an image [38]. We introduced three metrics based on gradient, entropy, and noise. The gradient-based metric ($M_{gradient}$) was used to evaluate the edge information. 

\begin{equation}
{\widetilde{g}}_i=\frac{1}{N_g}[\log{\lambda(}g_i-\gamma)+1]
\label{eq:eq7}
\end{equation}

\begin{equation}
N_g=\log{\left[\lambda\left(1-\gamma\right)+1\right]}
\label{eq:eq8}
\end{equation}

\begin{equation}
G_j=\sum_{i\in c_j}{\widetilde{g}}_i\ ,\ j=1,2,\ldots,N_c
\label{eq:eq9}
\end{equation}

\begin{equation}
M_{gradient}=\frac{K_GE\left(G\right)}{S\left(G\right)}
\label{eq:eq10}
\end{equation}

\noindent where ${\widetilde{g}}_i$ is the amount of gradient information at pixel $i$; $g_i$ is the gradient magnitude at pixel $i$, which is estimated using Sobel operator; $\lambda$ is the control parameter; $\gamma$ is the activation threshold; $N_g$ and $K_G$ are normalization factors; $G_j$ is the grid cell; $N_c$ is the number of grid cells; $E$ and $S$ are the mean and standard deviation operators, respectively.

The entropy-based metric was used to estimate the information of an image: 
\begin{equation}
M_{entropy}=-k_e\sum_{i=0}^{255}{P(i)\log_2{P(i)}}
\label{eq:eq11}
\end{equation}
where $P(i)$ is the probability of intensity level $i$ in a grayscale image; $k_e$ is a normalization factor. 

The noise-based metric is defined as follows:
\begin{equation}
M_{noise}=\sum_{j=1}^{3}{\sqrt{\frac{\pi}{2}}\frac{1}{N_p}\sum_{i} H\left(i\right)U\left(i\right)\left|I_j\ast K\right|(i)}
\label{eq:eq12}
\end{equation}
where $N_p$ is the number of pixels whose pixel value is 1 in $H.U$, where “$.$” is the elementwise multiplication operator; $H(i)$ is the homogeneous region mask for pixel $i$; $U(i)$ is the overexposed-underexposed region mask for pixel $i$; $I_j$ is the pixel value of the input image in channel $j$; and “$\ast$” is the convolution operator; $K$ is a noise estimation kernel that was used to evaluate the noise level of an image with Gaussian noise, as defined in Equation (\ref{eq:eq13}); the inhomogeneous, overexposed, and underexposed regions of images were removed by using a homogeneous mask and overexposed-underexposed mask defined in Equation (\ref{eq:eq14}) and Equation (\ref{eq:eq15}), since the noise estimation kernel overestimates noises in those regions.
\begin{equation}
K=\left[ \begin{array}{ccc}
1 & -2 & 1 \\ 
-2 & 4 & -2 \\ 
1 & -2 & 1 \end{array}
\right]
\label{eq:eq13}
\end{equation}
\begin{equation}
H(i)=\left\{ \begin{array}{c}
1\ \ \ g_i\le \delta  \\ 
0\ \ \ g_i>\delta  \end{array}
\right.
\label{eq:eq14}
\end{equation}
\begin{equation}
U(i)=\left\{ \begin{array}{c}
1\ \ \ {\tau }_l\le I_i\le {\tau }_u \\ 
0\ \ \ \ \ otherwise \end{array}
\right.
\label{eq:eq15}
\end{equation}
where $\delta$ is the adaptive threshold; $\tau_l$ and $\tau_u$ are the lower and upper bounds, respectively; and $I_i$ is the pixel value of pixel $i$. Finally, the image quality metric is derived by combining gradient, entropy, and noise-based metrics:
\begin{equation}
f\left(I\right)=AM_{gradient}+BM_{entropy}-CM_{noise}
\label{eq:eq16}
\end{equation}
where $A$, $B$, and $C$ are user parameters, which are set to 0.4, 0.6, and 0.6 according to [34]. A high $f\left(I\right)$ implies higher edge and texture details and lower noise. Therefore, it is promising to maximize the quality metric to obtain a well-exposed microscopy image.

\subsubsection{Particle swarm optimization}
With the noise-aware image quality metric, the particle swarm optimization algorithm was used to optimize the adjusted parameter, which is the only control parameter of the improved adaptive gamma correction method. The image quality metrics were used to define the objective function for optimization. We used 20 search agents that moved in the search space following Equation (\ref{eq:eq17}) to seek the optimal solutions in 30 iterations for five independent runs:
\begin{equation}
\phi_i^\prime=\phi_i+V_i^\prime
\label{eq:eq17}
\end{equation}
\begin{equation}
V_i^\prime=\omega V_i+c_1r_1\left({pbest}_i-\phi_i\right)+c_2r_2\left(gbest-\phi_i\right)
\label{eq:eq18}
\end{equation}
\begin{equation}
\omega=\left(\frac{itermax-iter}{itermax-1}\right)\left(\omega_{max}-\omega_{min}\right)+\omega_{min}
\label{eq:eq19}
\end{equation}
where $\phi_i^\prime$ and $\phi_i$ are the new and current positions of the $i$-th search agent; $V_i^\prime$ is the current velocity of the $i$-th agent; $V_i$ is the previous velocity of the $i$-th agent; ${pbest}_i$ is the best solution obtained by the $i$-th agent; $gbest$ is the global best solution; $\omega$ is the inertia weight; $\omega_{max}$ and $\omega_{min}$ are the maximum and minimum inertia weights, respectively; $c_1$ and $c_2$ are the acceleration coefficients; $r_1$ and $r_2$ are random numbers in the range of 0 to 1. 

\subsection{Image standardization and categorization}
\begin{figure}
  \centering
  \includegraphics[width=7cm]{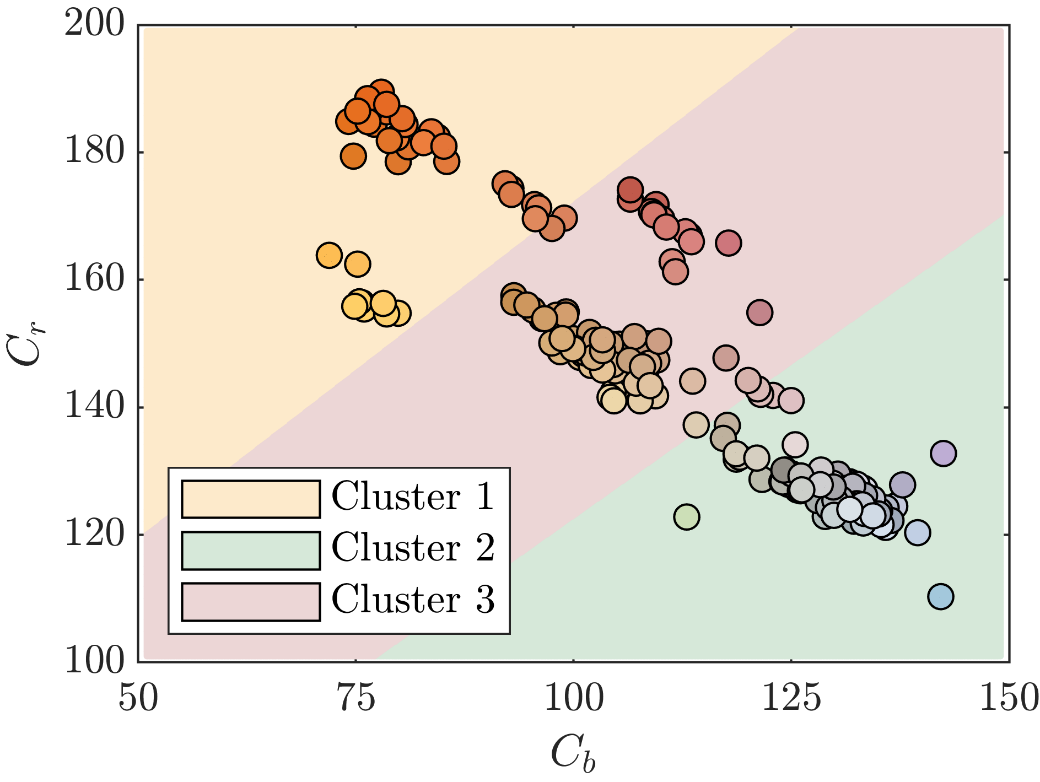}
  \caption{The geography of groups in the $C_b$ and $C_r$ space. Each dot represents a microscope image. The color of each dot represents the mean pixel intensity. The microscope images are grouped by k-means clustering based on the blue-difference and red-difference chroma components.}
  \label{fig:fig3}
\end{figure}

The images with different resolutions were resized to a standard size (256 × 256 pixels). Next, the resized images were normalized by dividing the pixel intensities in each channel by 255, so that the pixel intensity values were in the range of 0 to 1. With the different distributions of colors in the images, an unsupervised classification model was developed based k-means++ clustering with squared Euclidean distance to categorize the images into different groups according to the chroma channels ($C_b$ and $C_r$, see Figure \ref{fig:fig3}) [39]. 

\subsection{Semantic segmentation models}
A semantic segmentation model was developed based on an object-contextual representation network with high-resolution network [40, 41]. The model consisted of multiple convolutional neural networks for pixel-wise classification, and a convolutional neural network was trained for each class. A weak learning method was presented to enable the neural network to learn from new images while preserving the knowledge learned from previous images. Three methods were used for the imbalanced dataset: (1) A multi-class cross-entropy loss function was presented to mitigate the extent of imbalance. (2) An iterative stratification method was proposed to split the highly imbalanced dataset into train and test sets. (3) Image augmentation was performed to artificially enlarge the dataset. These methods are elaborated in the following subsections.

\subsubsection{Object-contextual representation}
The object-contextual representation network has six main steps (Figure \ref{fig:fig1}): 

\begin{enumerate}
\item Feature extraction: HRNetV2-W18 is used as the feature extractor by the neural network with 20 million parameters and pretrained using ImageNet for faster convergence [40]. 
\item Coarse segmentation: With the extracted features, a coarse segmentation map is generated for each class. The pixel value indicates the class of a pixel. The map is obtained according to the output of HRNetV2-W18 and a 1×1 convolution unit [40]. 
\item Object region representation: The object region representation for class $k$ is defined as:
\begin{equation}
f_k=\sum_{i\in I}{{\overline{m}}_{k,i}x_i}
\label{eq:eq20}
\end{equation}
\begin{equation}
{\overline{m}}_{k,i}=softmax\left(m_{k,i}\right)=\frac{e^{m_{k,i}}}{\sum_{j\in I} e^{m_{k,j}}}
\label{eq:eq21}
\end{equation}
where $f_k$ is the object region representation for class $k$; $x_i$ is the output of the backbone network for pixel $i$; ${\overline{m}}_{k,i}$ is the normalized value of $m_{k,i}$; $m_{k,i}$ is the value of pixel $i$ in coarse segmentation map for class $k$; the $softmax$ function is used to normalize the coarse segmentation maps. 

\item Pixel-region relation: The relation between pixels with each class is defined as:
\begin{equation}
\omega_{k,i}=\frac{e^{\psi\left(x_i\right).\psi(f_k)}}{\sum_{j=1}^{K}{\psi\left(x_i\right).\psi(f_j)}}
\label{eq:eq22}
\end{equation}
where $K$ is the number of classes; $\psi$ is the transformation function consisting of 1×1 convolution, batch normalization, and $ReLU$ activation function. 

\item Object contextual representation: The object contextual representation is computed by aggregating the pixel-region relation with all coarse segmentation maps:
\begin{equation}
y_i=\psi\left[\sum_{k=1}^{K}\omega_{k,i}\psi\left(f_k\right)\right]
\label{eq:eq23}
\end{equation}
where $y_i$ is the object contextual representation for pixel $i$; $f_j$ is the object region representation for class $k$; $K$ is the number of classes; and $\omega_{k,i}$ is the pixel-region relation for class $k$ and pixel $i$. 

\item Augmented representation: The final semantic segmentation map is obtained using the transformation function, the output of the backbone network, and object contextual representation: 
\begin{equation}
z_i=\psi([x_i\ y_i]T)
\label{eq:eq24}
\end{equation}
where $z_i$ is the final segmentation solution for pixel $i$; $x_i$ is the output of backbone network for pixel $i$; and $y_i$ is the object contextual representation for pixel $i$.

\end{enumerate}

\subsubsection{Weak learning}
Weak learning was proposed to address two challenges in training of the convolutional neural networks: (1) The dataset is limited and becomes smaller when it is divided into different groups. (2) It is time-consuming to repeatedly train the convolutional neural network when new images are added to the dataset.

Weak learning was performed in two steps: (1) A convolutional neural network was trained using the whole dataset. (2) Transfer learning was performed. The neural network was retained for each group of images while the learning rate was near zero. The learning rate is a hyperparameter of convolutional neural network and controls the changes of the weights in training. Since the learning rate was low, the change of weights was small, so the knowledge learned from previous images was reserved. Meanwhile, the convolutional neural network learned new images and slightly modified the weights to achieve high performance for the new images.

\subsubsection{Multi-class cross-entropy loss}
Cross-entropy was applied to obtain the discrepancy between the predicted distribution mask $P\left(x\right)$ and the true distribution mask $D\left(x\right)$:
\begin{equation}
CE\left(P,D\right)=\sum_{x\epsilon \mathrm{\Omega }}{D\left(x\right)\mathrm{log}\mathrm{}(P\left(x\right))}
\label{eq:eq25}
\end{equation}
where $\Omega$ is the mask region. The multi-class cross-entropy loss function is expressed as:

\begin{equation}
{Loss}_C=\sum_{i=1}^{K}\sum_{x\epsilon\Omega} {d_{x,i}\mathrm{log}\mathrm{}(p_{x,i})}
\label{eq:eq26}
\end{equation}
where $K$ is the number of classes; $d_{x,i}$ is the true one-hot distribution probability of pixel $x$ on class $i$; and $p_{x,i}$ is the predicted probability for pixel $x$ and class $i$. 

Since the annotated pixels of under-represented classes were scarce, the effect of the under-represented classes on the loss function was limited. Thus, the neural network tends to maximize the prediction accuracy for over-represented classes while ignoring the under-presented classes. This research assigned sample weights to mitigate the representation problem:
\begin{equation}
{Loss}_W=\sum_{i=1}^{K}\sum_{x\epsilon\Omega}{{w_id}_{x,i}\mathrm{log}\mathrm{}(p_{x,i})}
\label{eq:eq27}
\end{equation}
where $w_i$ is the sample weight for class $i$. 

There was no definitive advice on how to set the sample weights. We proposed to calculate the sample weights using the following equation:
\begin{equation}
w_i={(\frac{1}{\mu_i})}^\beta
\label{eq:eq28}
\end{equation}
where $\mu_i$ is the class weight for class $i$; and $\beta$ is an adjustment factor. The adjustment factor was optimized by the particle swarm optimization to minimize the image quality metric.

\subsubsection{Iterative stratification}
When the dataset is small and highly imbalanced, it is inappropriate to randomly split the dataset into train and test sets, because it is possible that under-represented classes are missing in one of the sets and thus the trained model fails to represent the whole dataset. We proposed to employ an iterative stratification method for multi-label data [35]. The whole dataset was divided into more than two subsets while the class weights of the sets were the same. 

\subsubsection{Image augmentation}
Augmentation was performed to increase the dataset size and improve the generalizability of the semantic segmentation model [44]. The training dataset was enlarged using four strategies: (1) Resizing: The images were resized from 256×256 pixels to 320×256 pixels. (2) Random cropping: The images were randomly cropped to 256×256 pixels. (3) Random flipping: The images were randomly flipped horizontally and vertically with a probability of 50\%. (4) Photometric distortion: The brightness, contrast, saturation, and hue of images were randomly modified with a probability of 50\%. 

\subsection{Performance metrics}
Six performance metrics were used to evaluate the semantic segmentation model:
\begin{equation}
Pixel\ accuracy=\frac{\sum_{k=1}^{K}{(TP_k+TN_k)}}{\sum_{k=1}^{K}{(TP_k+FP_k+TN_k+FN_k)}}
\label{eq:eq29}
\end{equation}
\begin{equation}
Mean\ accuracy=\frac{1}{K}\sum_{k=1}^{K}\left(\frac{TP_k+TN_k}{TP_k+FP_k+TN_k+FN_k}\right)
\label{eq:eq30}
\end{equation}
\begin{equation}
mIoU=\frac{1}{K}\sum_{k=1}^{K}\left(\frac{TP_k}{TP_k+FP_k+FN_k}\right)
\label{eq:eq31}
\end{equation}
\begin{equation}
Precision=\frac{1}{K}\sum_{k=1}^{K}\left(\frac{TP_k}{TP_k+FP_k}\right)
\label{eq:eq32}
\end{equation}
\begin{equation}
Recall=\frac{1}{K}\sum_{k=1}^{K}\left(\frac{TP_k}{TP_k+FN_k}\right)
\label{eq:eq33}
\end{equation}
\begin{equation}
F1=2\times\frac{Precision}{Precision+Recall}
\label{eq:eq34}
\end{equation}
where $TP_k$ and $TN_k$ are respectively the true positives and true negatives corresponding to class $k$; $FP_k$ and $FN_k$ are respectively the false positives and false negatives corresponding to class $k$; $K$ is the number of classes; and $mIoU$ is mean intersection-over-union.

\section{Results}
\begin{figure}%
    \centering
    \subfloat{{\includegraphics[width=3.5cm]{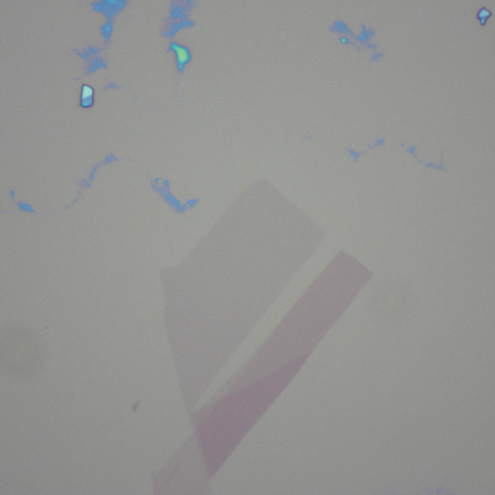} }}%
    \qquad    
    \subfloat{{\includegraphics[width=3.5cm]{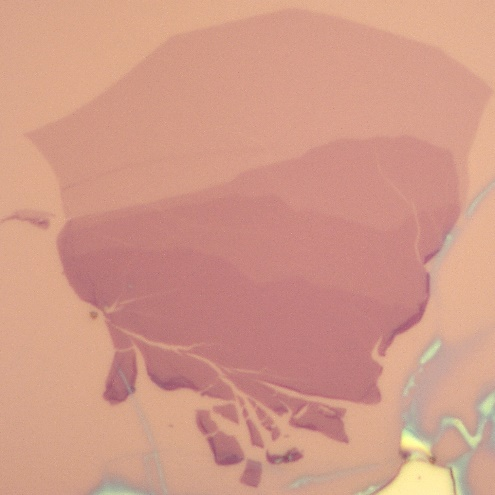} }}%
    \qquad
    \subfloat{{\includegraphics[width=3.5cm]{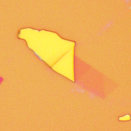} }}%
    \qquad
    \subfloat{{\includegraphics[width=3.5cm]{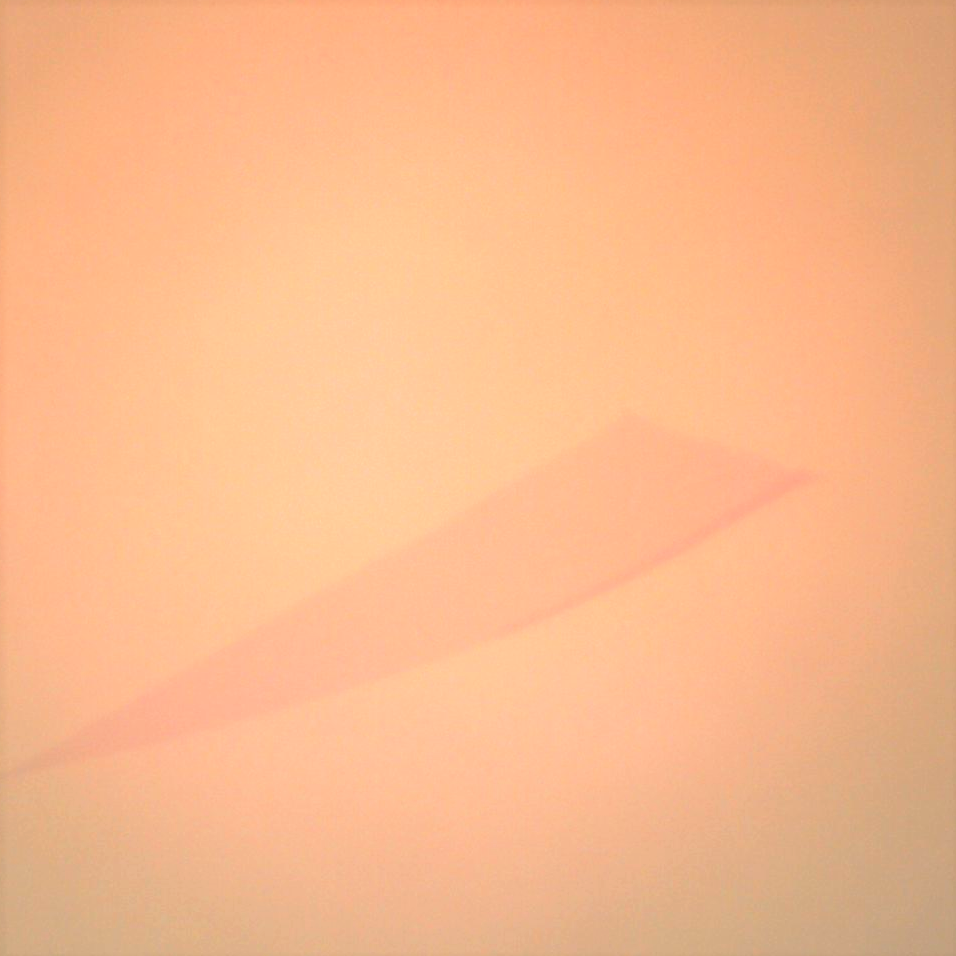} }}%
    \qquad
    \subfloat{{\includegraphics[width=3.5cm]{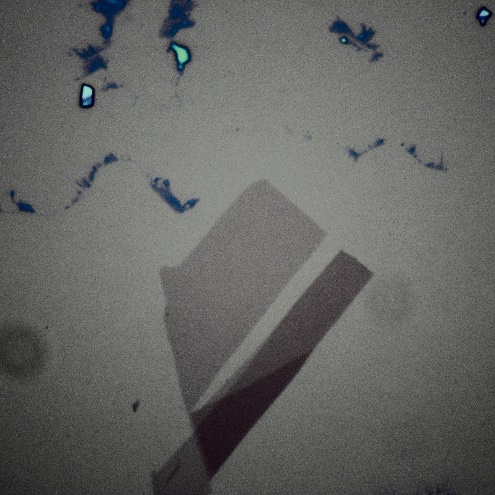} }}%
    \qquad    
    \subfloat{{\includegraphics[width=3.5cm]{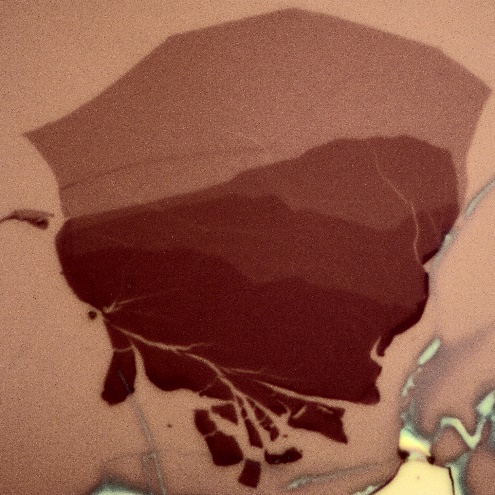} }}%
    \qquad
    \subfloat{{\includegraphics[width=3.5cm]{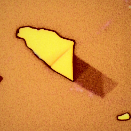} }}%
    \qquad
    \subfloat{{\includegraphics[width=3.5cm]{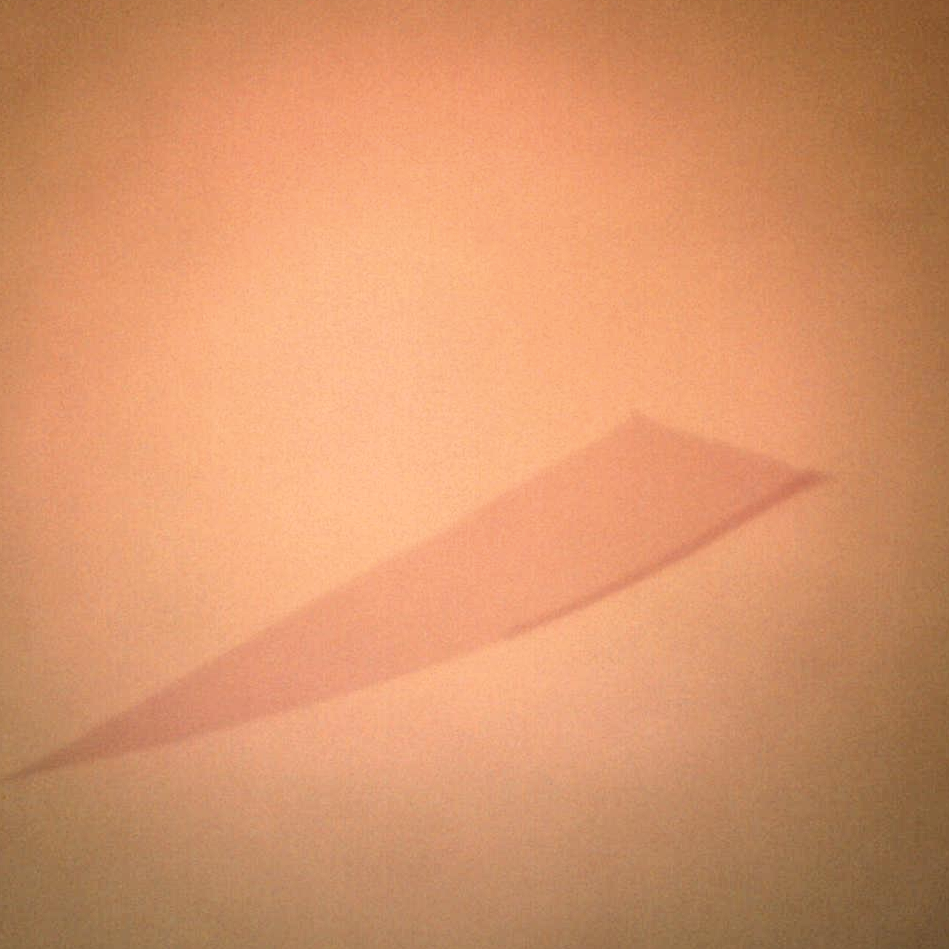} }}%
    \qquad

    \caption{The effect of optimized adaptive gamma correction on the microscope images. The upper row are the original images, and the lower row of images are the enhanced images by the optimized adaptive gamma correction.}%
    \label{fig:fig4}
\end{figure}

\begin{figure}%
    \centering
    \subfloat[]{{\includegraphics[width=7.6cm]{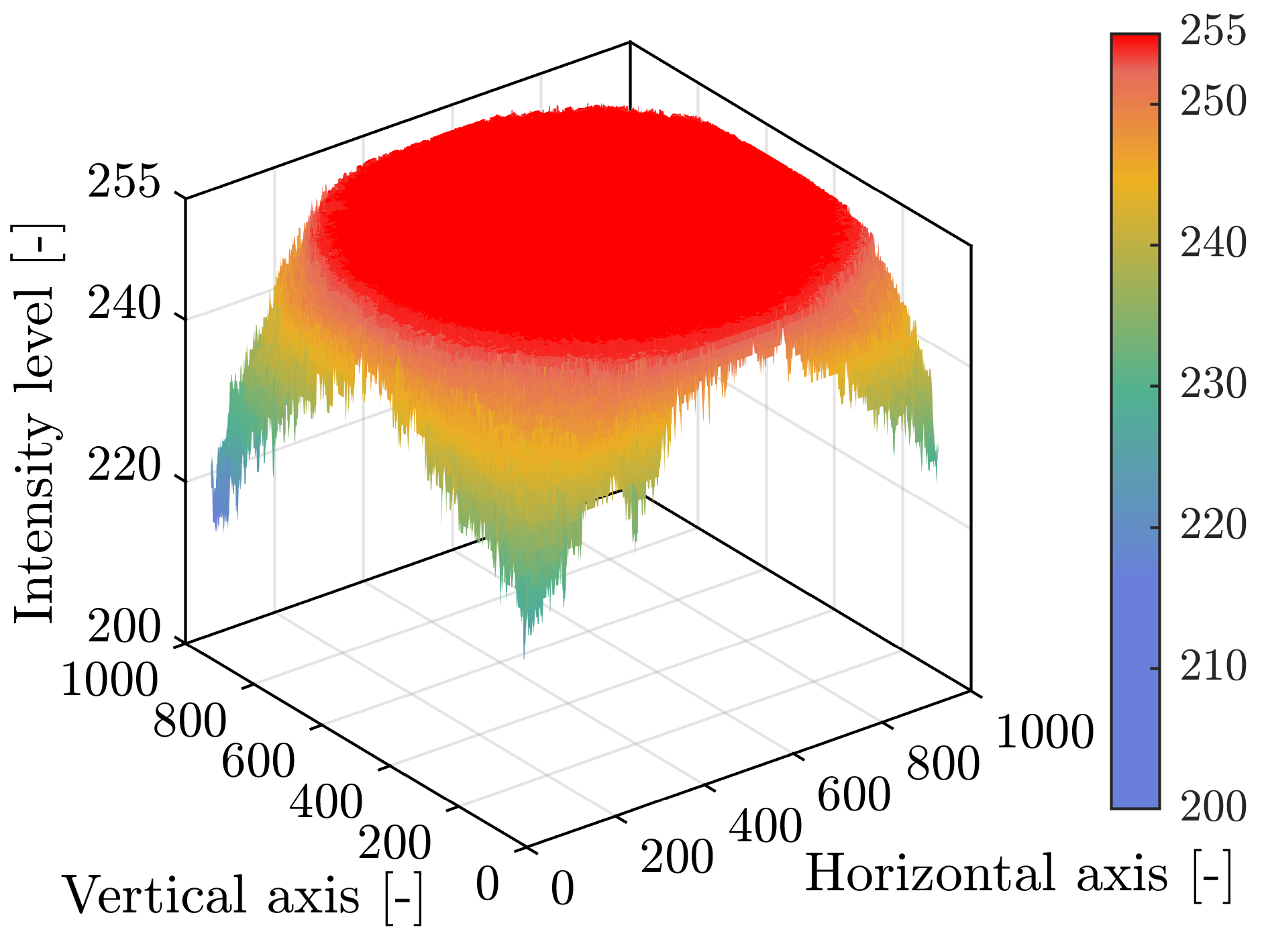} }}%
    \qquad    
    \subfloat[]{{\includegraphics[width=7.6cm]{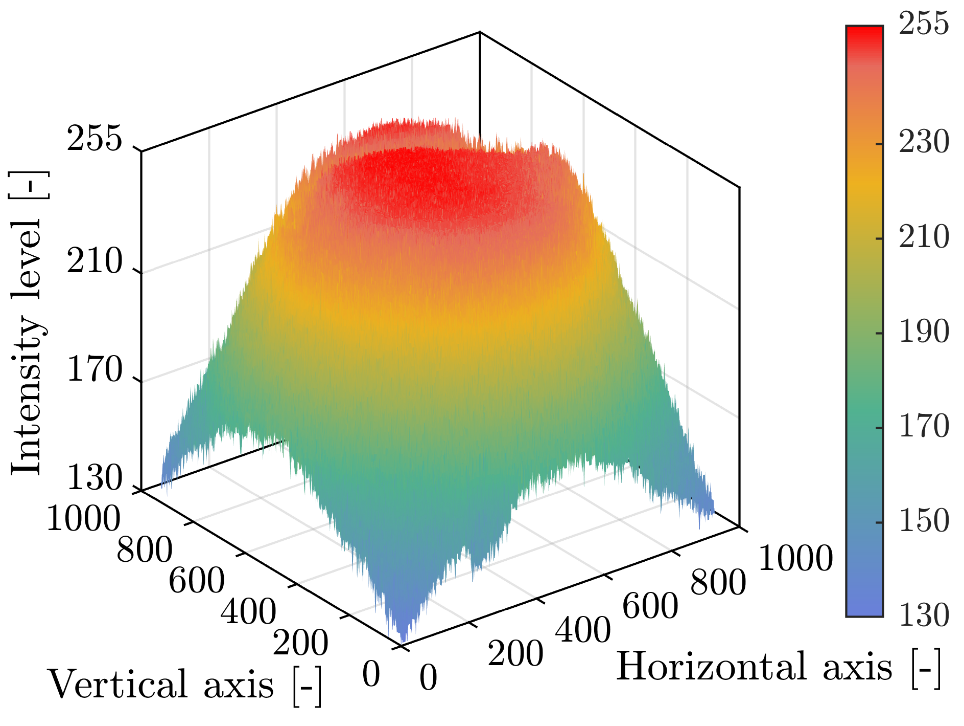} }}%
    \qquad
    \subfloat[]{{\includegraphics[width=7.6cm]{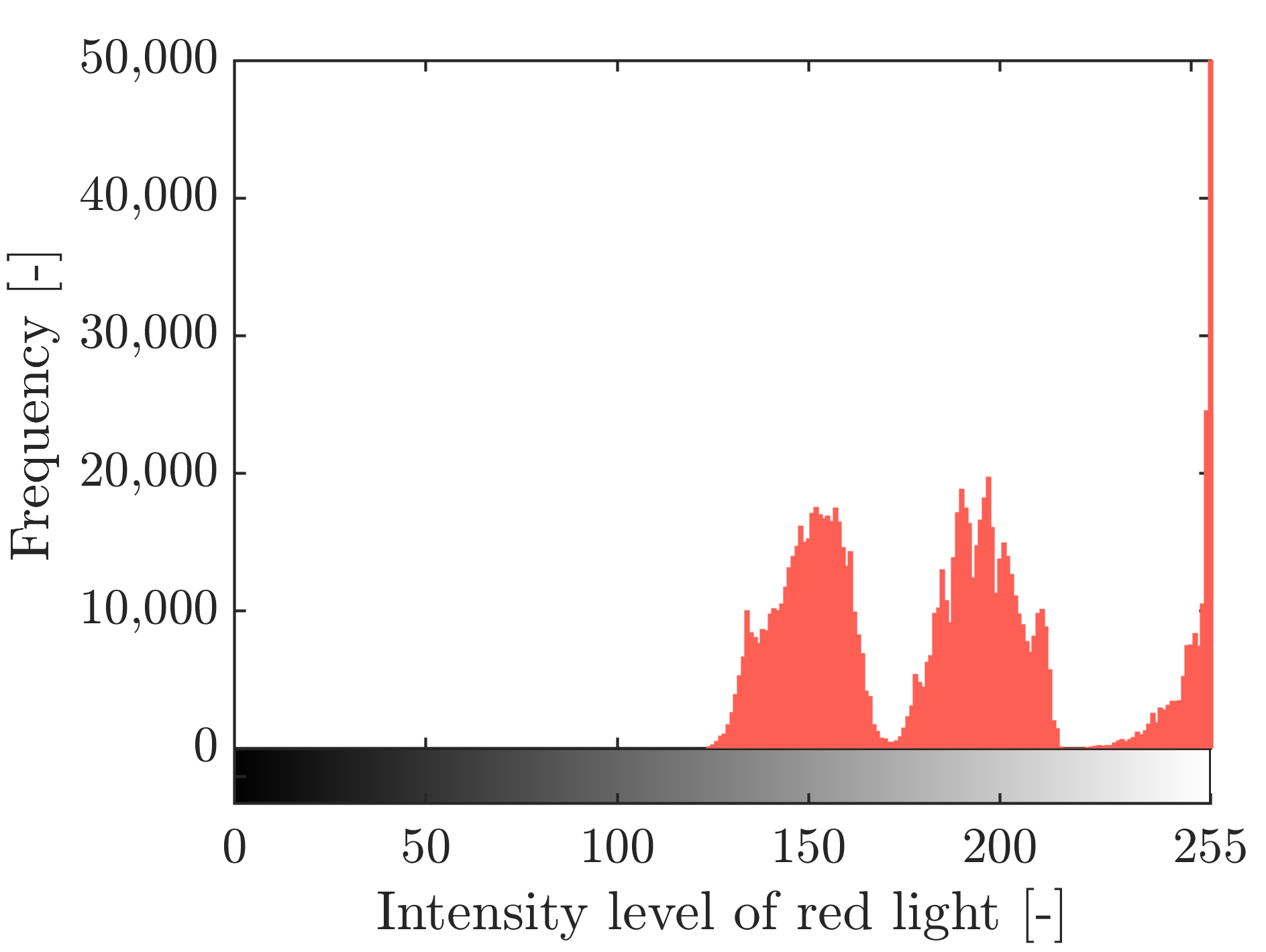} }}%
    \qquad
    \subfloat[]{{\includegraphics[width=7.6cm]{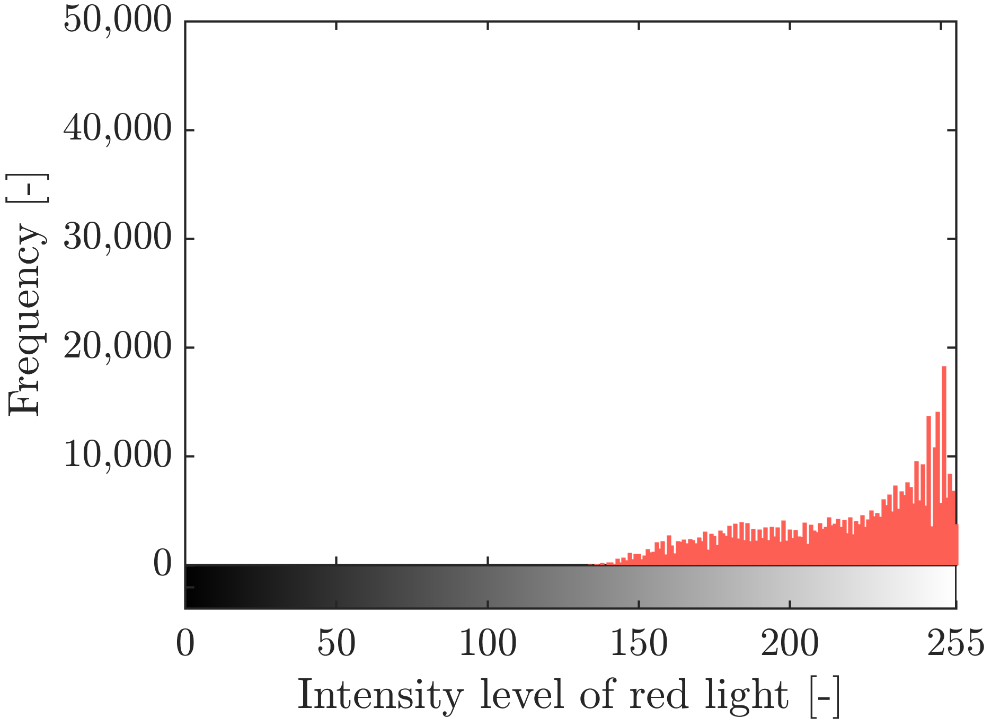} }}%
    \qquad
    \subfloat[]{{\includegraphics[width=4.8cm]{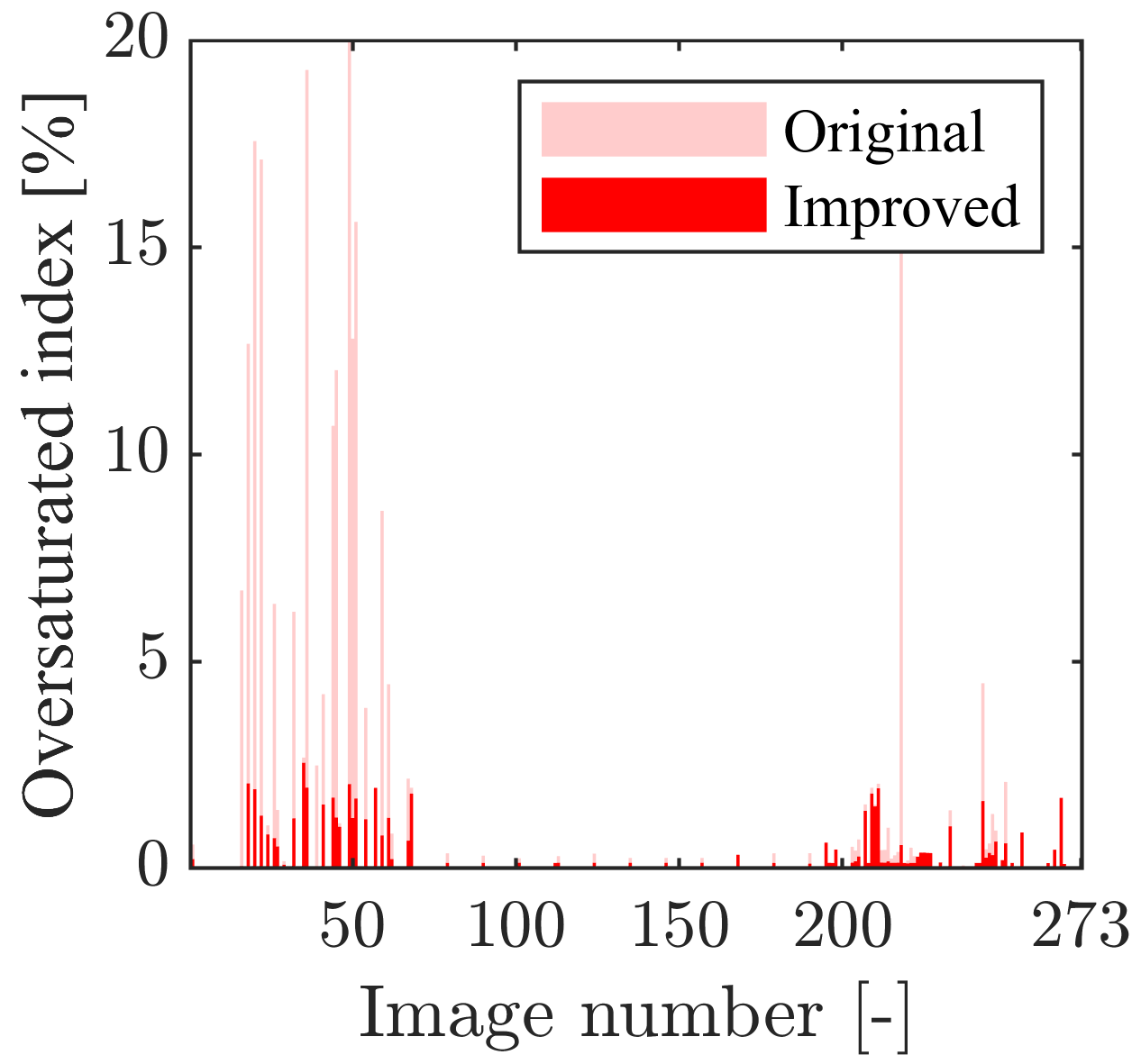} }}%
    \qquad    
    \subfloat[]{{\includegraphics[width=4.8cm]{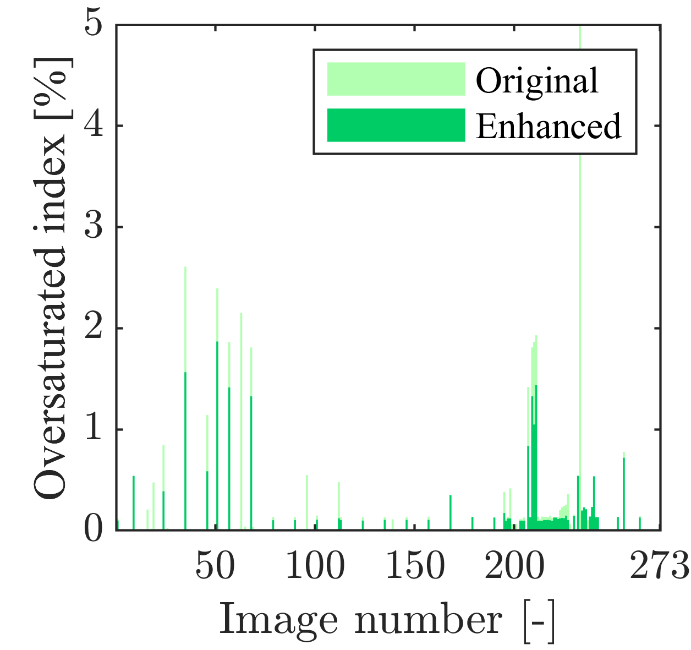} }}%
    \qquad
    \subfloat[]{{\includegraphics[width=4.8cm]{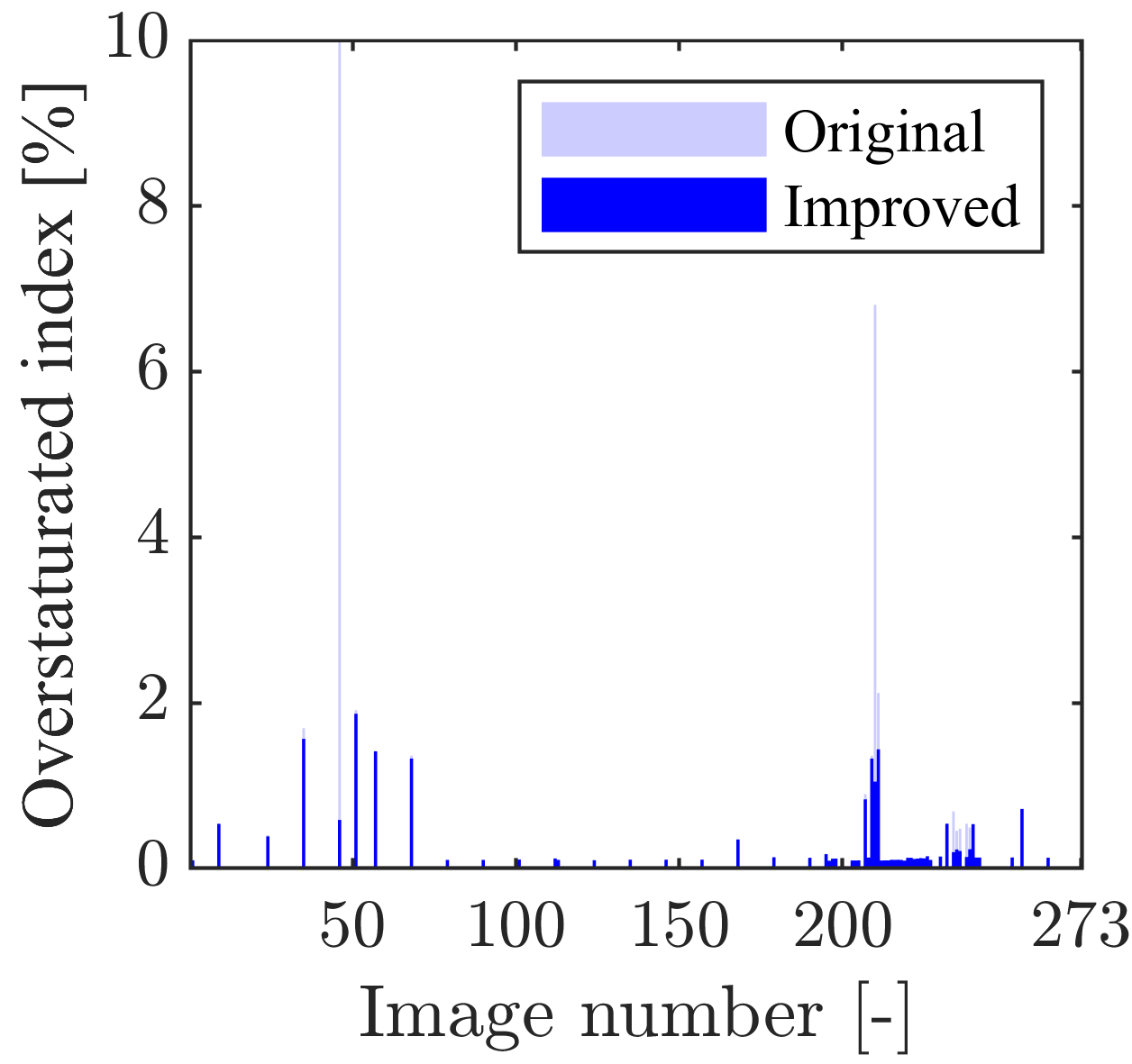} }}%
    \qquad

    \caption{An investigation on an overexposed image: (a) the original image, (b) the improved image, (c) the pixel intensity histogram of the red channel of the original image, (d) the pixel intensity histogram of the red channel of the improved image, (e) oversaturated index of the red channel, (f) oversaturated index of the green channel, and (g) oversaturated index of the blue channel. In (a) and (b), pixels with intensity levels over 253 represent oversaturation and are shown in red color.}%
    \label{fig:fig5}
\end{figure}

Four pairs of representative images before and after applying the optimized adaptive gamma correction are compared in Figure \ref{fig:fig4}. The comparison shows that the optimized adaptive gamma correction method is capable of improving the visibility of the microscope images of graphene regardless of the background color and brightness. The texture and edges of the improved images become easier to distinguish from the background. The optimal adjusted parameter obtained by the particle swarm optimization performed to minimize the image quality metric was 0.561. The lowest image quality metric value was 1.48. 
The pixel intensity levels of the fourth pair of images are compared in Figure \ref{fig:fig5}. The comparison shows that the oversaturation and oversaturated regions are greatly reduced by impeding the optimized adaptive gamma correction. An index defined by Equation (\ref{eq:eq35}) is proposed to indicate the effect of the optimized adaptive gamma correction for improvement of the images:
\begin{equation}
\delta=\frac{A_o}{A_T}
\label{eq:eq35}
\end{equation}
where $\delta$ is the oversaturated index; $A_o$ is the number of pixels with an intensity level over 253; and $A_T$ is the total number of pixels. The oversaturated indices of original and modified images for the red, green, and blue channels are highly reduced as shown in  Figure \ref{fig:fig5} (e) to Figure \ref{fig:fig5} (g).

The proposed method (HRNet+OCR) was compared with three state-of-the-art convolutional neural networks, which are UNet++ [45], PSPNet [46], and DeepLabv3+ [47]. The proposed models were trained by stochastic gradient descent. The learning rate was 0.1. The momentum was 0.9. The weight decay was 0.0005. The batch size was eight. The maximum number of iterations was 10,000. The cross-entropy loss was used as the loss function. The performance metrics of the different methods are compared in Table \ref{tab:table2}. The proposed method shows the highest accuracy. 

\begin{table}
 \caption{Results of the performance metrics}
  \centering
  \begin{threeparttable}
  \begin{tabular}{lllllllll}
    \toprule
Methods & Pixel accuracy & Mean accuracy & F1 & mIoU & Precision & Recall \\
    \midrule
    The existing methods \\
UNet++ & 96.7\% & 61.7\% & 56.0 & 56.4 & 60.9 & 51.9\% \\
PSPNet & 95.5\% & 64.6\% & 56.0 & 56.5 & 59.5 & 52.8\% \\ 
DeepLabv3+ & 96.9\% & 70.9\% & 62.8 & 58.5 & 66.7 & 59.3\% \\ 
The proposed methods\\
Baseline (HRNet+OCR) & 97.1\% & 71.7\% & 65.9 & 59.0 & 67.9 & 63.9\% \\ 
Baseline+$Loss{}_{W}^1$ & 98.3\% & 73.6\% & 72.9 & 61.5 & 77.2 & 69.1\% \\ 
Baseline+$Loss{}_{W}$+OAGC$^2$ & 98.8\% & 75.2\% & 87.2 & 66.7 & 89.7 & 85.1\% \\ 
Baseline+$Loss{}_{W}$+OAGC+WL$^3$ & 99.0\% & 81.6\% & 91.8 & 71.7 & 93.5 & 90.2\% \\

    \bottomrule
  \end{tabular}
  \begin{tablenotes}
  \item[1] "$Loss{}_{w}$" denotes weighted cross-entropy loss.
  \item[2] "OAGC" denotes optimized adaptive gamma correction.
  \item[3] "WL" denotes weak learning.
  \end{tablenotes}
  \end{threeparttable}
  \label{tab:table2}

\end{table}

\begin{figure}%
    \centering
    \subfloat[]{{\includegraphics[width=14cm]{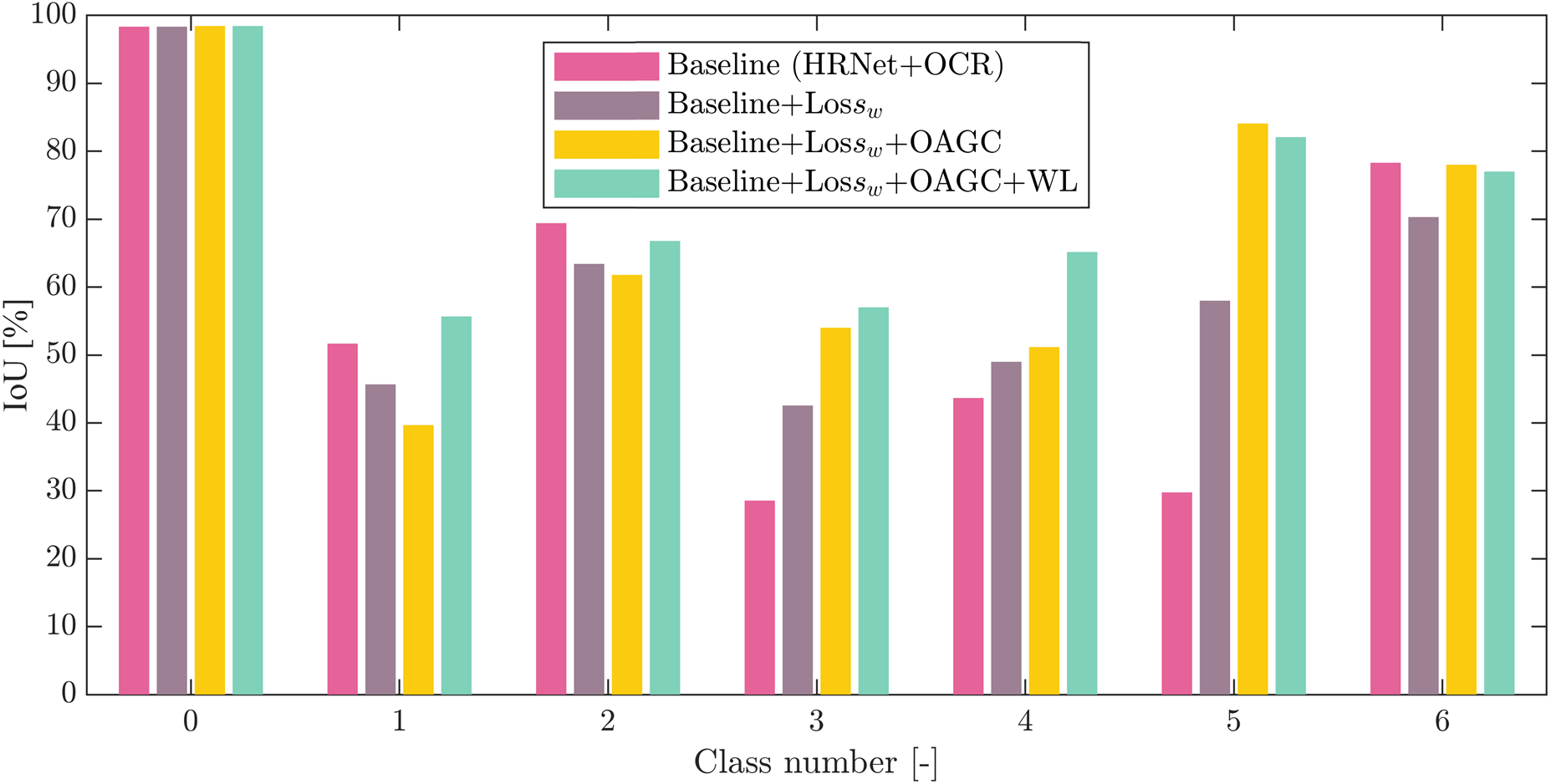} }}%
    \qquad
    \subfloat[]{{\includegraphics[width=14cm]{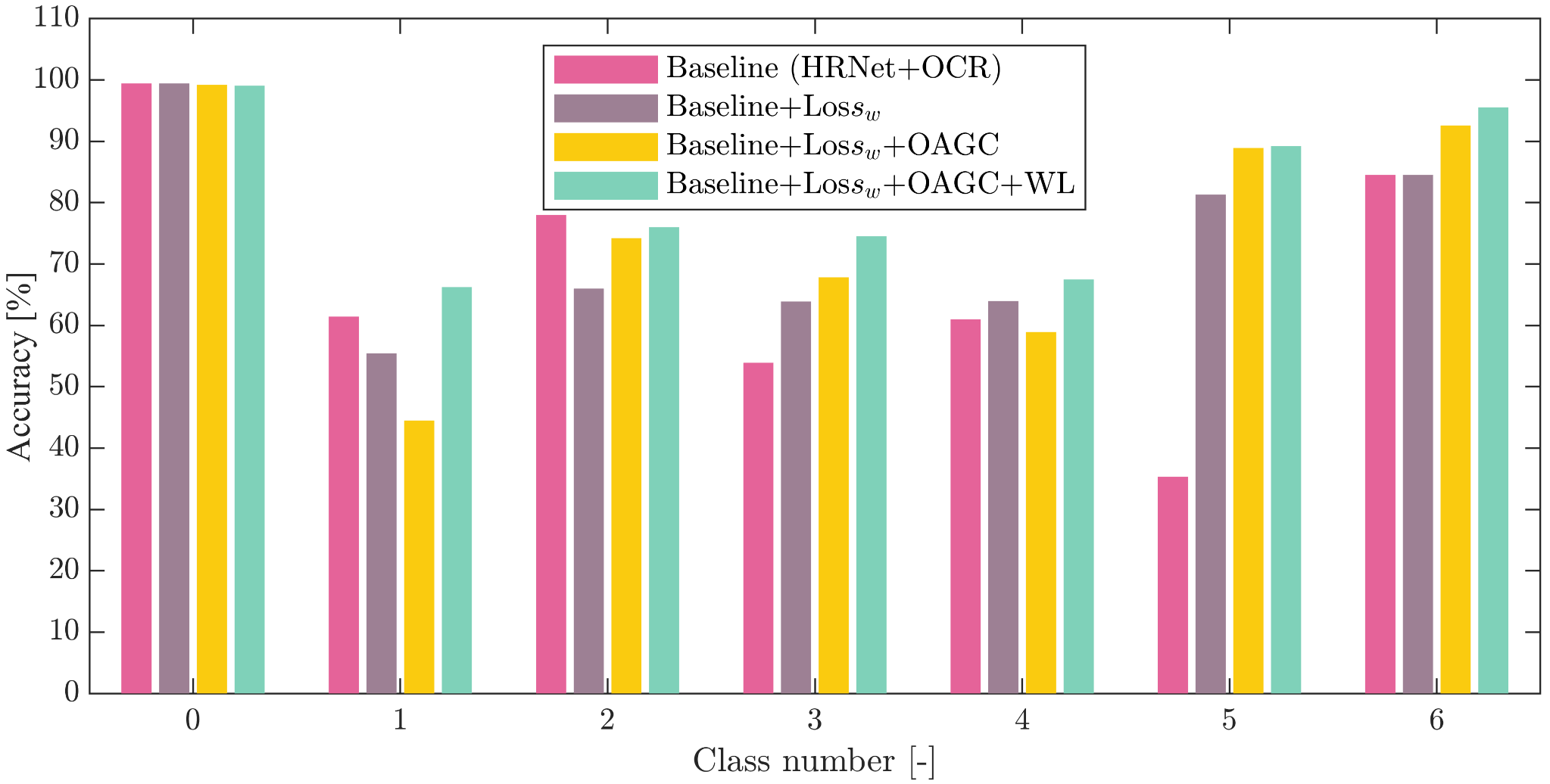} }}%
    \qquad

    \caption{The class-wise performance of the applied segmentation methods: (a) Mean IoU (mIoU), (b) accuracy. $Loss{}_{W}$, OAGC, WL stand for the weighted cross-entropy loss, optimized adaptive gamma correction, and weak learning, respectively.}%
    \label{fig:fig6}
\end{figure}

\begin{figure}%
\centering

\begin{tabular}{ p{3.7cm} p{3.7cm} p{3.7cm} p{3.7cm}}
      \hspace{0.35cm} Microscope image &  \hspace{0.35cm} Ground-truth label & \hspace{0.6cm} Predicted result & \hspace{1.2cm} Overlay \\

\end{tabular}

\begin{tabular}{@{}c@{}}

    \subfloat{\frame{\includegraphics[width=3.7cm,height=2.62cm]{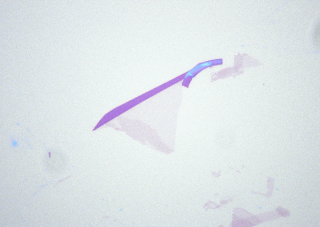} }}\hspace{1.1em}%
    \subfloat{\frame{\includegraphics[width=3.7cm,height=2.62cm]{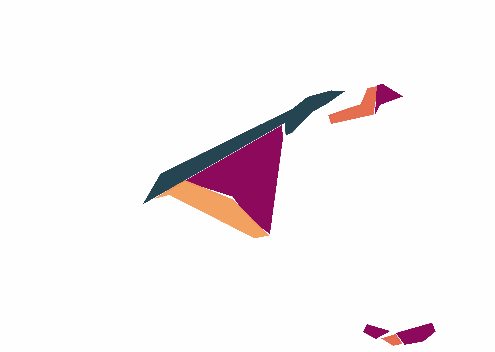} }}\hspace{1.1em}%
    \subfloat{\frame{\includegraphics[width=3.7cm,height=2.62cm]{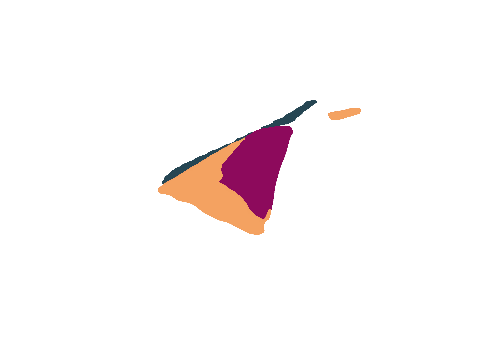} }}\hspace{1.1em}%
    \subfloat{\frame{\includegraphics[width=3.7cm,height=2.62cm]{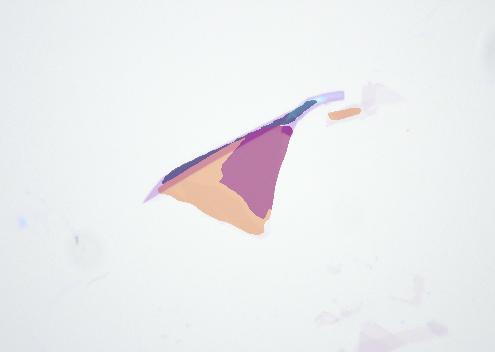} }}\hspace{1.1em}%
\end{tabular}\qquad

\begin{tabular}{@{}c@{}}
    \subfloat{\frame{\includegraphics[width=3.7cm,height=2.62cm]{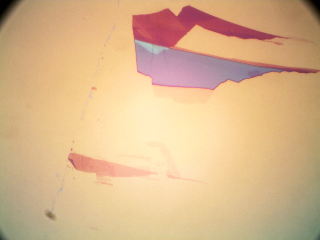} }}\hspace{1.1em}%
    \subfloat{\frame{\includegraphics[width=3.7cm,height=2.62cm]{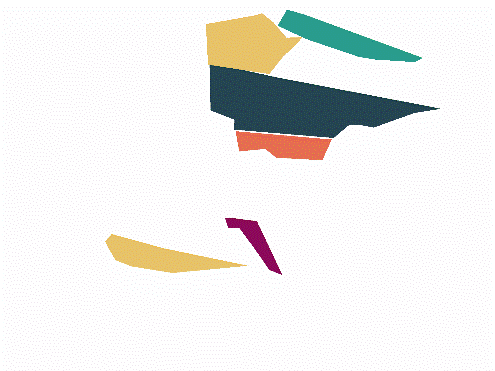} }}\hspace{1.1em}%
    \subfloat{\frame{\includegraphics[width=3.7cm,height=2.62cm]{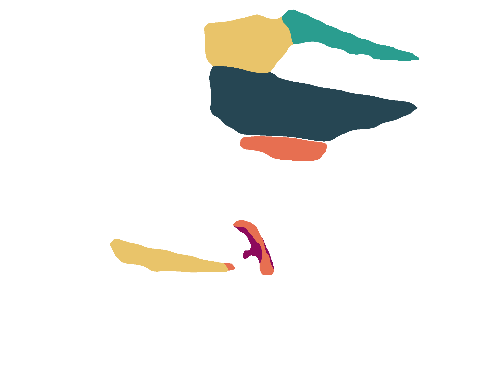} }}\hspace{1.1em}%
    \subfloat{\frame{\includegraphics[width=3.7cm,height=2.62cm]{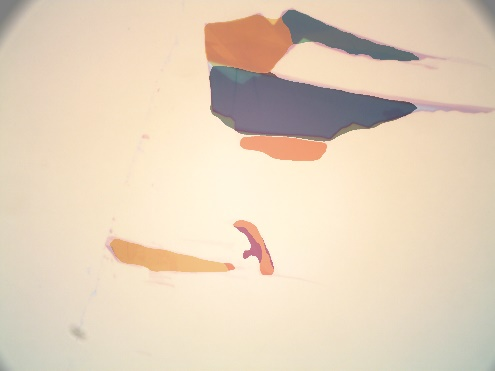} }}\hspace{1.1em}%
\end{tabular}\qquad
\begin{tabular}{@{}c@{}}

    \subfloat{\frame{\includegraphics[width=3.7cm,height=2.62cm]{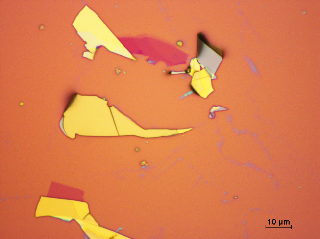} }}\hspace{1.1em}%
    \subfloat{\frame{\includegraphics[width=3.7cm,height=2.62cm]{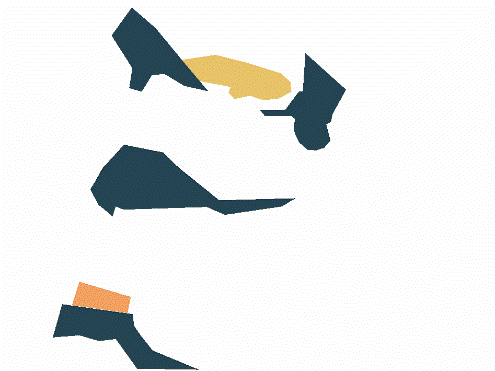} }}\hspace{1.1em}%
    \subfloat{\frame{\includegraphics[width=3.7cm,height=2.62cm]{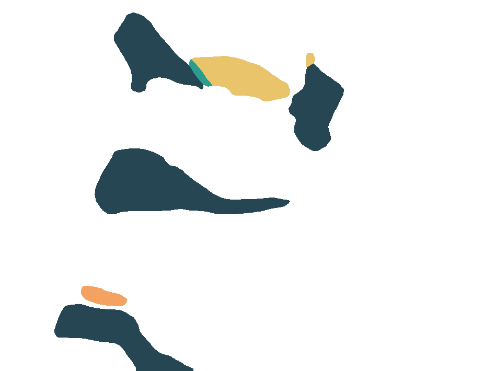} }}\hspace{1.1em}%
    \subfloat{\frame{\includegraphics[width=3.7cm,height=2.62cm]{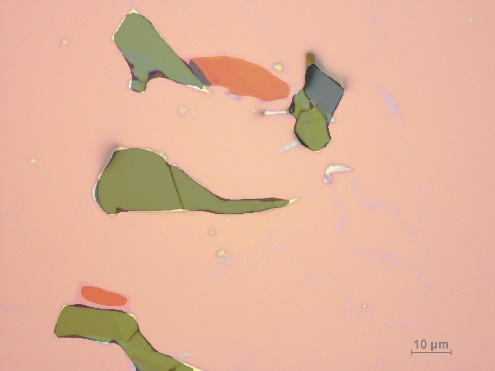} }}\hspace{1.1em}%
\end{tabular}\qquad
\begin{tabular}{@{}c@{}}
    \subfloat{\frame{\includegraphics[width=3.7cm,height=2.62cm]{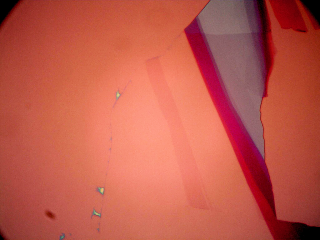} }}\hspace{1.1em}%
    \subfloat{\frame{\includegraphics[width=3.7cm,height=2.62cm]{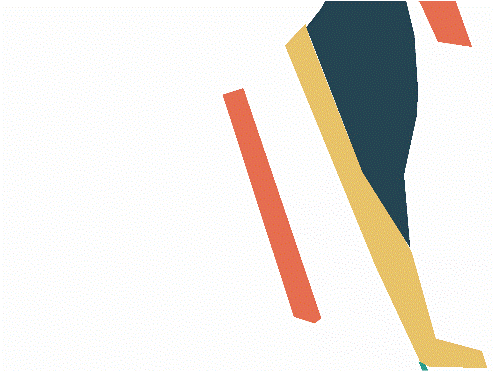} }}\hspace{1.1em}%
    \subfloat{\frame{\includegraphics[width=3.7cm,height=2.62cm]{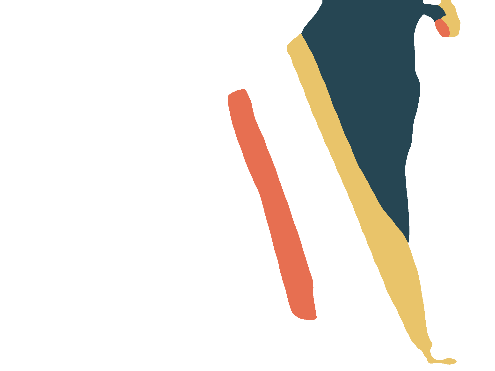} }}\hspace{1.1em}%
    \subfloat{\frame{\includegraphics[width=3.7cm,height=2.62cm]{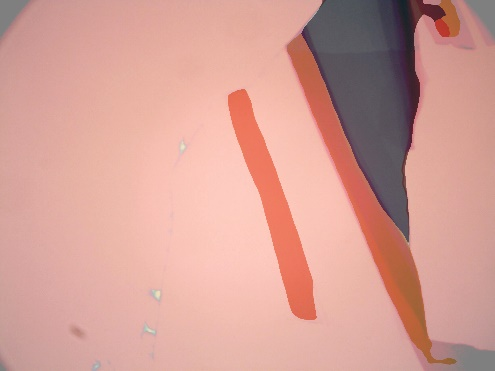} }}\hspace{1.1em}%
\end{tabular}\qquad
\begin{tabular}{@{}c@{}}
    \subfloat{\frame{\includegraphics[width=3.7cm,height=2.62cm]{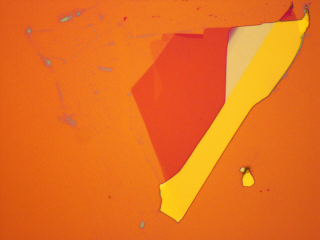} }}\hspace{1.1em}%
    \subfloat{\frame{\includegraphics[width=3.7cm,height=2.62cm]{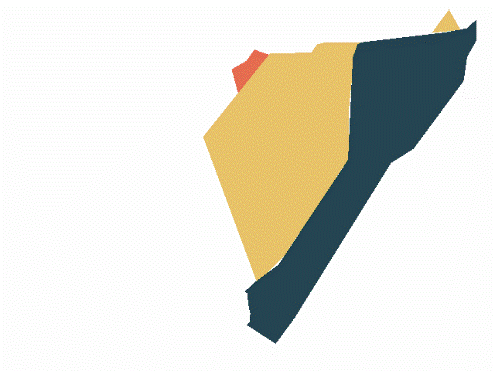} }}\hspace{1.1em}%
    \subfloat{\frame{\includegraphics[width=3.7cm,height=2.62cm]{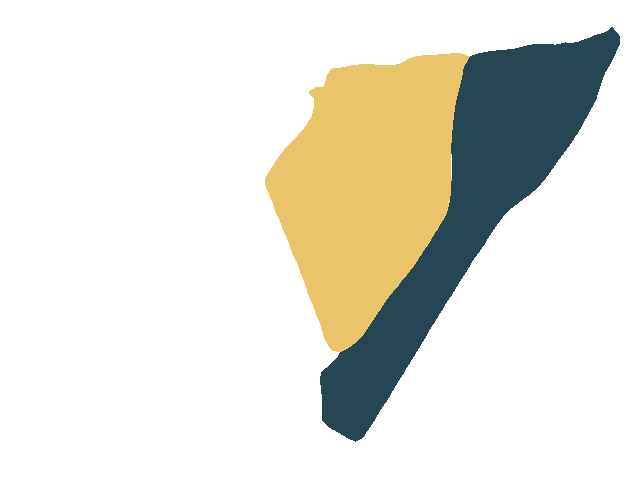} }}\hspace{1.1em}%
    \subfloat{\frame{\includegraphics[width=3.7cm,height=2.62cm]{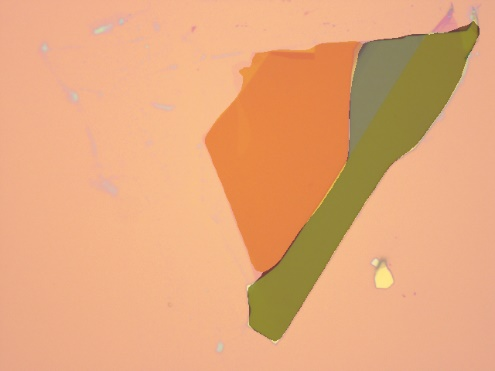} }}\hspace{1.1em}%
\end{tabular}\qquad
\begin{tabular}{@{}c@{}}

    \subfloat{\frame{\includegraphics[width=3.7cm,height=2.62cm]{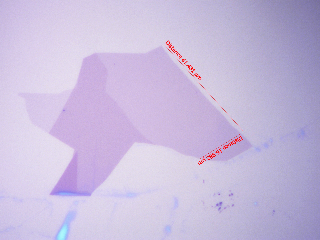} }}\hspace{1.1em}%
    \subfloat{\frame{\includegraphics[width=3.7cm,height=2.62cm]{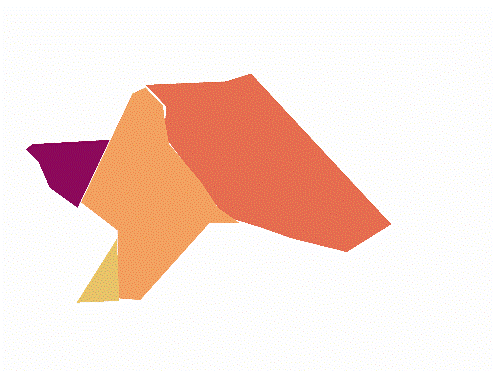} }}\hspace{1.1em}%
    \subfloat{\frame{\includegraphics[width=3.7cm,height=2.62cm]{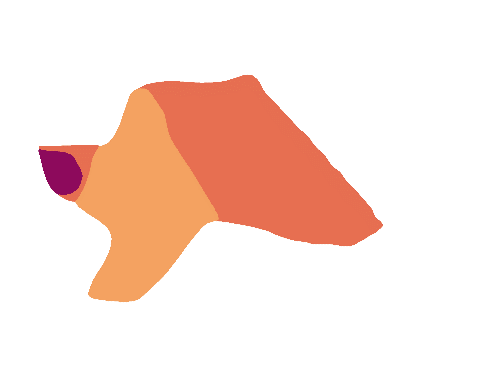} }}\hspace{1.1em}%
    \subfloat{\frame{\includegraphics[width=3.7cm,height=2.62cm]{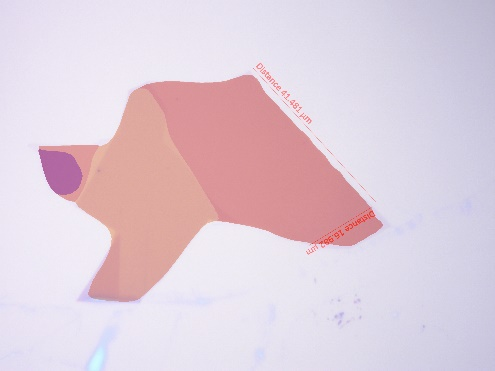} }}\hspace{1.1em}%
\end{tabular}\qquad
     \subfloat{{\includegraphics[width=13cm]{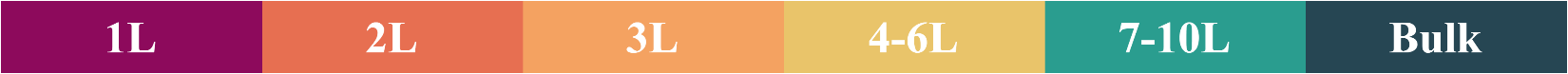} }}\par

    \caption{Representative optical microscope images and the predicted results using the final model (HRNet+OCR+$Loss{}_{W}$+OAGC+WL).}%
    \label{fig:fig7}
\end{figure}

Further, four HRNet+OCR models were compared to test the effects of the weighted cross-entropy loss, optimized adaptive gamma correction, and weak learning on the performance. The results show that the accuracy is greatly improved by applying the weighted cross-entropy loss, optimized adaptive gamma correction, and weak learning. The pixel accuracy of the final model (HRNet+OCR+$Loss{}_{W}$+OAGC+WL) is higher than 99\%. The class-wise intersection-over-union (IoU) and accuracy of the HRNet+OCR models are shown in Figure \ref{fig:fig6}. The IoU and accuracy of the highly under-represented classes 1L, 3L, and 6-10L are greatly increased by the weighted cross-entropy loss, optimized adaptive gamma correction, and weak learning. The minimum IoU of the final model is higher than 56\%. Figure \ref{fig:fig7} shows representative images in the test set and the corresponding prediction results. 

\section{Conclusions}
For the first time, we developed a deep learning method for automatic identification and classification of exfoliated graphene flakes from optical microscope images with various microscope settings. Novel methods such as optimized adaptive gamma correction are presented to improve the microscopy images and address the challenges associated with the imbalanced classes, limited dataset size, different background conditions, and different resolutions of the microscopy images. The presented machine learning framework based on object-contextual representations is promising for automatically identifying 2D material flakes and classifying the thickness of the flakes based on microscope images. The pixel accuracy of the trained model is higher than 99\%, outperforming the state-of-the-art deep convolutional neural networks DeepLabv3+, PSPNet, and U-Net ++. The proposed method is robust to the background colors of microscope images because the proposed machine learning method can automatically identify the background color of the images and categorize the images into appropriate groups. The proposed method is also robust to the brightness and resolution of images because the optimized adaptive gamma correction method can effectively improve the quality of overexposed microscopy images. The presented machine learning framework is capable of learning characterization from new images while preserving the knowledge learned from old images. Weak learning method enables us to retrain convolutional neural networks while preserving the learned knowledge by the network. The method is useful especially when the training set is small but there is a convolutional neural network trained on a similar task. The presented method can handle imbalanced datasets. The IoU of different classes is in the range of 57\% and 99\%, and the mIoU of all the classes is 59\%. This methodology shows that image augmentation, iterative stratification, and weighted cross-entropy loss are able to significantly enhance the predictive performance of convolutional neural networks on under-represented classes and successfully identify the optical images with various backgrounds.

\section*{Acknowledgement}
This research was funded by National Science Foundation (grant number CMMI-2046407).

%Bibliography
\bibliography{references}  
[1] B. Radisavljevic, A. Radenovic, J. Brivio, V. Giacometti, A. Kis, Single-layer MoS2 transistors, Nature Nanotechnology, 6 (2011) 147-150.

\noindent [2] B. Huang, G. Clark, E. Navarro-Moratalla, D.R. Klein, R. Cheng, K.L. Seyler, D. Zhong, E. Schmidgall, M.A. McGuire, D.H. Cobden, W. Yao, D. Xiao, P. Jarillo-Herrero, X. Xu, Layer-dependent ferromagnetism in a van der Waals crystal down to the monolayer limit, Nature, 546 (2017) 270-273.

\noindent [3] C. Gong, L. Li, Z. Li, H. Ji, A. Stern, Y. Xia, T. Cao, W. Bao, C. Wang, Y. Wang, Z.Q. Qiu, R.J. Cava, S.G. Louie, J. Xia, X. Zhang, Discovery of intrinsic ferromagnetism in two-dimensional van der Waals crystals, Nature, 546 (2017) 265-269.

\noindent [4] X. Xu, W. Yao, D. Xiao, T.F. Heinz, Spin and pseudospins in layered transition metal dichalcogenides, Nature Physics, 10 (2014) 343-350.

\noindent [5] X. Xi, Z. Wang, W. Zhao, J.-H. Park, K.T. Law, H. Berger, L. Forró, J. Shan, K.F. Mak, Ising pairing in superconducting NbSe2 atomic layers, Nature Physics, 12 (2016) 139-143.

\noindent [6] X. Zhang, D. Sun, Y. Li, G.-H. Lee, X. Cui, D. Chenet, Y. You, T. F. Heinz, and J. Hone, Measurement of lateral and interfacial thermal conductivity of single- and bilayer MoS2 and MoSe2 using optothermal Raman technique, ACS Applied Materials \& Interfaces, 7 (2015) 25923-25929.

\noindent [7] Y. Li, F. Ye, J. Xu, W. Zhang, P. Feng, and X. Zhang, Gate-tuned temperature in a hexagonal boron nitride-encapsulated 2D semiconductor devices, IEEE Transactions on Electronic Devices, 65 (2018) 4068-4072.

\noindent [8] E. Easy, Y. Gao, Y. Wang, D. Yan, S.M. Goushehgir, E.-H. Yang, B. Xu, X. Zhang, Experimental and computational investigation of layer-dependent thermal conductivities and interfacial thermal conductance of one- to three-layer WSe2, ACS Applied Materials \& Interfaces, 13 (2021) 13063-13071.

\noindent [9] E. Easy, Y. Gao, Y. Wang, D. Yan, S.M. Goushehgir , E.-H. Yang , B. Xu, X. Zhang , Thermal conductivities and interfacial thermal conductance of 1- to 3-layer WSe2, arXiv:2011.00730, (2020).

\noindent [10] F. Ye, Q. Liu, B. Xu, P.X.L. Feng, X Zhang, Very high interfacial thermal conductance in fully hBN-encapsulated MoS2 van der Waals heterostructure, ArXiv:2102.05239, (2021).

\noindent [11] Y. Li, A. Chernikov, X. Zhang, A. Rigosi, H. M. Hill, A. M. van der Zande, D. A. Chenet, E.-M. Shih, J. Hone, and T. F. Heinz, Measurement of the optical dielectric function of monolayer transition-metal dichalcogenides: MoS2, MoSe2, WS2, WSe2, Physical Review B, 90 (2014) 205422.

\noindent [12] X. Zhang, Characterization of layer number of two-dimensional transition metal diselenide semiconducting devices using Si-peak analysis, Advances in Materials Science and Engineering, 2019 (2019) 7865698.

\noindent [13] K.S. Novoselov, A.K. Geim, S.V. Morozov, D. Jiang, Y. Zhang, S.V. Dubonos, I.V. Grigorieva, A.A. Firsov, Electric field effect in atomically thin carbon films, Science, 306 (2004) 666.

\noindent [14] K.S. Novoselov, D. Jiang, F. Schedin, T.J. Booth, V.V. Khotkevich, S.V. Morozov, A.K. Geim, Two-dimensional atomic crystals, Proceedings of the National Academy of Sciences of the United States of America, 102 (2005) 10451.

\noindent [15] H. Li, J. Wu, X. Huang, G. Lu, J. Yang, X. Lu, Q. Xiong, H. Zhang, Rapid and reliable thickness identification of two-dimensional nanosheets using optical microscopy, ACS Nano, 7 (2013) 10344-10353.

\noindent [16] X. Lin, Z. Si, W. Fu, J. Yang, S. Guo, Y. Cao, J. Zhang, X. Wang, P. Liu, K. Jiang, W. Zhao, Intelligent identification of two-dimensional nanostructures by machine-learning optical microscopy, Nano Research, 11 (2018) 6316-6324.

\noindent [17] S. Masubuchi, T. Machida, Classifying optical microscope images of exfoliated graphene flakes by data-driven machine learning, npj 2D Materials and Applications, 3 (2019) 4.

\noindent [18] Z.H. Ni, H.M. Wang, J. Kasim, H.M. Fan, T. Yu, Y.H. Wu, Y.P. Feng, Z.X. Shen, Graphene thickness determination using reflection and contrast spectroscopy, Nano Letters, 7 (2007) 2758-2763.

\noindent [19] C.M. Nolen, G. Denina, D. Teweldebrhan, B. Bhanu, A.A. Balandin, High-throughput large-area automated identification and quality control of graphene and few-layer graphene films, ACS Nano, 5 (2011) 914-922.

\noindent [20] P. Blake, E.W. Hill, A.H. Castro Neto, K.S. Novoselov, D. Jiang, R. Yang, T.J. Booth, A.K. Geim, Making graphene visible, Applied Physics Letters, 91 (2007) 063124.

\noindent [21] E. Shelhamer, J. Long, T. Darrell, Fully convolutional networks for semantic segmentation, IEEE Transactions on Pattern Analysis and Machine Intelligence, 39 (2017) 640-651.

\noindent [22] L.-C. Chen, G. Papandreou, I. Kokkinos, K. Murphy, A.L. Yuille, DeepLab: Semantic image segmentation with deep convolutional nets, atrous convolution, and fully connected CRFs, IEEE Transactions on Pattern Analysis and Machine Intelligence, 40 (2018) 834-848.

\noindent [23] O. Ronneberger, P. Fischer, T. Brox, U-Net: convolutional networks for biomedical image segmentation, in: Medical Image Computing and Computer-Assisted Intervention – MICCAI 2015, Springer International Publishing, Cham, 2015, pp. 234-241.

\noindent [24] V. Badrinarayanan, A. Kendall, R. Cipolla, SegNet: A deep convolutional encoder-decoder architecture for image segmentation, IEEE Transactions on Pattern Analysis and Machine Intelligence, 39 (2017) 2481-2495.

\noindent [25] S. Masubuchi, M. Morimoto, S. Morikawa, M. Onodera, Y. Asakawa, K. Watanabe, T. Taniguchi, T. Machida, Autonomous robotic searching and assembly of two-dimensional crystals to build van der Waals superlattices, Nature Communications, 9 (2018) 1413.

\noindent [26] B. Han, Y. Lin, Y. Yang, N. Mao, W. Li, H. Wang, K. Yasuda, X. Wang, V. Fatemi, L. Zhou, J.I.J. Wang, Q. Ma, Y. Cao, D. Rodan-Legrain, Y.-Q. Bie, E. Navarro-Moratalla, D. Klein, D. MacNeill, S. Wu, H. Kitadai, X. Ling, P. Jarillo-Herrero, J. Kong, J. Yin, T. Palacios, Deep-learning-enabled fast optical identification and characterization of 2d materials, Advanced Materials, 32 (2020) 2000953.

\noindent [27] Y. Saito, K. Shin, K. Terayama, S. Desai, M. Onga, Y. Nakagawa, Y.M. Itahashi, Y. Iwasa, M. Yamada, K. Tsuda, Deep-learning-based quality filtering of mechanically exfoliated 2D crystals, npj Computational Materials, 5 (2019) 124.

\noindent [28] E. Greplova, C. Gold, B. Kratochwil, T. Davatz, R. Pisoni, A. Kurzmann, P. Rickhaus, M.H. Fischer, T. Ihn, S.D. Huber, Fully automated identification of two-dimensional material samples, Physical Review Applied, 13 (2020) 064017.

\noindent [29] S. Masubuchi, E. Watanabe, Y. Seo, S. Okazaki, T. Sasagawa, K. Watanabe, T. Taniguchi, T. Machida, Deep-learning-based image segmentation integrated with optical microscopy for automatically searching for two-dimensional materials, npj 2D Materials and Applications, 4 (2020) 3.

\noindent [30] J. Kirkpatrick, R. Pascanu, N. Rabinowitz, J. Veness, G. Desjardins, A.A. Rusu, K. Milan, J. Quan, T. Ramalho, A. Grabska-Barwinska, D. Hassabis, Overcoming catastrophic forgetting in neural networks, Proceedings of the National Academy of Sciences, 114 (2017) 3521-3526.

\noindent [31] W. Chen, X. Mao, H. Ma, Low-contrast microscopic image enhancement based on multi-technology fusion, 2010 IEEE International Conference on Intelligent Computing and Intelligent Systems, IEEE, 2010, pp. 891-895.

\noindent [32] F.W. Leong, M. Brady, J.O.D. McGee, Correction of uneven illumination (vignetting) in digital microscopy images, Journal of Clinical Pathology, 56 (2003) 619-621.
\noindent [33] Z. Wu, Y. Gao, L. Li, J. Xue, Y. Li, Semantic segmentation of high-resolution remote sensing images using fully convolutional network with adaptive threshold, Connection Science, 31 (2019) 169-184.

\noindent [34] G. Cao, L. Huang, H. Tian, X. Huang, Y. Wang, R. Zhi, Contrast enhancement of brightness-distorted images by improved adaptive gamma correction, Computers \& Electrical Engineering, 2018 (66) 569-582.

\noindent [35] K. Sechidis, G. Tsoumakas, I. Vlahavas, On the stratification of multi-label data, Joint European Conference on Machine Learning and Knowledge Discovery in Databases, Springer, 2011, pp. 145-158.

\noindent [37] R. Poli, J. Kennedy, T. Blackwell, Particle swarm optimization. Swarm Intelligence, 2007, 1(1), pp.33-57.

\noindent [38] U. Shin, J. Park, G. Shim, F. Rameau, I.S. Kweon, Camera exposure control for robust robot vision with noise-aware image quality assessment, 2019 IEEE/RSJ International Conference on Intelligent Robots and Systems (IROS).

\noindent [39] D. Arthur, S. Vassilvitskii, k-means++: The advantages of careful seeding, Stanford, 2006.

\noindent [40] Y. Yuan, X. Chen, J. Wang, Object-contextual representations for semantic segmentation, in: A. Vedaldi, H. Bischof, T. Brox, J.-M. Frahm (Eds.) Computer Vision – ECCV 2020, Springer International Publishing, Cham, 2020, pp. 173-190.

\noindent [41] J. Wang, K. Sun, T. Cheng, B. Jiang, C. Deng, Y. Zhao, D. Liu, Y. Mu, M. Tan, X. Wang, W. Liu, Deep high-resolution representation learning for visual recognition, IEEE Transactions on Pattern Analysis and Machine Intelligence, 2021, 43(10), 3349-3364.

\noindent [42] Z. Tian, L. Liu, B. Fei, Deep convolutional neural network for prostate MR segmentation, Medical Imaging 2017: Image-Guided Procedures, Robotic Interventions, and Modeling, International Society for Optics and Photonics, 2017, pp. 101351.

\noindent [43] X. Zhou, C. Yao, H. Wen, Y. Wang, S. Zhou, W. He, J. Liang, EAST: An efficient and accurate scene text detector, 2017 IEEE Conference on Computer Vision and Pattern Recognition (CVPR), 2017, pp. 2642-2651.

\noindent [44] C. Shorten, T.M. Khoshgoftaar, A survey on image data augmentation for deep learning, Journal of Big Data, 6 (2019) 60.

\noindent [45] Z. Zhou, M.M. Rahman Siddiquee, N. Tajbakhsh, J. Liang, UNet++: A nested U-Net architecture for medical image segmentation, in: Deep Learning in Medical Image Analysis and Multimodal Learning for Clinical Decision Support, Springer International Publishing, Cham, 2018, pp. 3-11.

\noindent [46] H. Zhao, J. Shi, X. Qi, X. Wang, J. Jia, Pyramid scene parsing network, Proceedings of the IEEE Conference on Computer Vision and Pattern Recognition, 2017, pp. 2881-2890.

\noindent [47] L.-C. Chen, Y. Zhu, G. Papandreou, F. Schroff, H. Adam, Encoder-decoder with atrous separable convolution for semantic image segmentation,  Proceedings of the European Conference on Computer Vision (ECCV), 2018, pp. 801-818.

\end{document}